\documentclass[a4paper,fleqn]{cas-dc}

\usepackage[numbers,longnamesfirst]{natbib}

\usepackage{framed,multirow}

\usepackage{amssymb}
\usepackage{latexsym}

\usepackage{url}
\usepackage{xcolor}

\usepackage{times}
\usepackage{epsfig}
\usepackage{graphicx}
\usepackage{amsmath}
\usepackage{amssymb}
\usepackage{fixltx2e}
\usepackage{amsfonts}
\usepackage{listings}
\usepackage{algorithmic}
\usepackage{booktabs} 
\usepackage{multirow}
\usepackage{amssymb} 
\usepackage{gensymb}
\usepackage{mathtools}
\usepackage{adjustbox}
\usepackage{nameref}

\usepackage[labelfont=bf]{caption}

\definecolor{DarkGreen}{rgb}{0.2,0.5,0.2} 

\makeatletter
\AtBeginDocument{\def\@citecolor{DarkGreen}}
\makeatother

\usepackage{arydshln}
\newcommand{\etal}{\textit{et al}.}

\def\tsc#1{\csdef{#1}{\textsc{\lowercase{#1}}\xspace}}
\tsc{WGM}
\tsc{QE}
\tsc{EP}
\tsc{PMS}
\tsc{BEC}
\tsc{DE}

\begin{document}
\let\WriteBookmarks\relax
\def\floatpagepagefraction{1}
\def\textpagefraction{.001}

\shorttitle{}

\shortauthors{Fard et~al.}

\title [mode = title]{Linguistic-Based Mild Cognitive Impairment Detection Using Informative Loss}                      


%

\author[1]{Ali Pourramezan Fard}[type=editor,
                        orcid=0000-0002-3807-0798]
\ead{Ali.PourramezanFard@du.edu}
\credit{Conceptualization of this study, Methodology, Implementation, Experimental Results and Analysis.}

\author[1,2]{Mohammad H. Mahoor}[type=editor,
                        orcid=0000-0001-8923-4660]
\ead{mmahoor@du.edu}
\cormark[1]
\credit{Evaluation of the study, Methodology, and Editing.}


\author[1,4]{Muath Alsuhaibani}[type=editor,
                        orcid=0009-0007-5164-012X]
\ead{muath.alsuhaibani@du.edu}
\cormark[0]
\credit{Analysis of the related research and Editing.}


\author[3]{Hiroko H. Dodge}[type=editor,
                        orcid=0000-0001-7290-8307]
\ead{hdodge@mgh.harvard.edu}
\cormark[0]


\cortext[cor1]{Corresponding author}
\credit{Providing Dataset and Editing.}

\affiliation[1]{organization={Ritchie School of Engineering and Computer Science},
    addressline={University of Denver}, 
    city={Denver},
    postcode={CO 80208}, 
    country={USA}}

\affiliation[2]{organization={DreamFace Technologies LLC},
    city={Centennial},
    postcode={CO 8011}, 
    country={USA}}
    
\affiliation[3]{organization={Department of Neurology, Massachusetts General Hospital},
    addressline={Harvard Medical School}, 
    city={Boston},
    postcode={MA 02114}, 
    country={USA}}

\affiliation[4]{organization={Department of Electrical Engineering, Prince Sattam Bin Abdulaziz University},
            city={Al-Kharj},
            postcode={11942}, 
            country={Saudi Arabia}}

\begin{abstract}
This paper presents a deep learning method using Natural Language Processing (NLP) techniques, to distinguish between Mild Cognitive Impairment (MCI) and Normal Cognitive (NC) conditions in older adults. We propose a framework that analyzes transcripts generated from video interviews collected within the I-CONECT study project, a randomized controlled trial aimed at improving cognitive functions through video chats. Our proposed NLP framework consists of two Transformer-based modules, namely Sentence Embedding (SE) and Sentence Cross Attention (SCA). First, the SE module captures contextual relationships between words within each sentence. Subsequently, the SCA module extracts temporal features from a sequence of sentences. This feature is then used by a Multi-Layer Perceptron (MLP) for the classification of subjects into MCI or NC. To build a robust model, we propose a novel loss function, called InfoLoss, that considers the reduction in entropy by observing each sequence of sentences to ultimately enhance the classification accuracy. The results of our comprehensive model evaluation using the I-CONECT dataset show that our framework can distinguish between MCI and NC with an average area under the curve of 84.75\%. 

\end{abstract}




\maketitle

\section{Introduction}


Alzheimer’s disease (AD) and AD-related dementia (AD/ADRD), the sixth leading cause of death in the United States since 20221~\cite{fs}, is a progressive neurological disorder primarily affecting cognitive function, and leading to memory loss~\cite{alzWhatAlzheimers}. About 6.7 million Americans, age 65 and older, are diagnosed with AD/ADRD in 2023~\cite{fs}, and unfortunately, this number is estimated to reach 13.8 million by 2060~\cite{fs}.

MCI is a transitional state between normal aging and the noticeable cognitive decline and difficulties seen in the early stages of ADRD~\cite{Petersen2014}. While memory loss is a distinctive characteristic of MCI, it can also impact other cognitive domains such as language, attention, and problem-solving~\cite{gilles2022age}.  

Diagnosing MCI typically involves a comprehensive evaluation by a healthcare professional, often a neurologist or geriatrician, using a combination of clinical assessments, medical history, cognitive testing, and, in some cases, medical imaging. These methods are costly and time-consuming. Thus, alternative methods applying Artificial Intelligence (AI) for affordable and ecologically valid assessments, can potentially lead to substantial time and financial savings. Impairments in language skills are present in a wide variety of neurodegenerative diseases~\cite{boschi2017connected}. This is especially noticeable in cases of dementia as aggravations in linguistic capabilities are frequently observed in both the early and advanced stages of the disease~\cite{boschi2017connected, calza2021linguistic}. As AD/ADRD advances, linguistic capabilities undergo a widespread decline in speech comprehension and verbal expression and ultimately become limited only to repetitive and stereotypical responses~\cite{ferris2013language}. Recently, there has been a growing interest in using NLP techniques, particularly for the early detection of MCI~\cite{roark2011spoken, toledo2018analysis, santos2017enriching, clarke2021comparison, yeung2021correlating, hussein2022natural, penfold2022development, amini2023automated}.

In this study, we introduce a deep learning-based framework designed to distinguish between MCI and Normal Cognitive (NC) conditions using NLP techniques. Our approach involves analyzing the transcriptions of conversations conducted between older adults participating in interviews (referred to as participants)) and the corresponding interviewers. We utilize the data provided by the Internet-Based Conversational Engagement Clinical Trial (I-CONECT) project~\cite{yu2021internet} (Clinicaltrials.gov \#: NCT02871921). The I-CONECT Study is a randomized controlled trial behavioral intervention (conversational interactions), with the primary aim to enhance cognitive functions by providing frequent social interactions through video chats. Throughout six months, older individuals (aged 75 and older) in the experimental
group participated in semi-structured conversations lasting 30 minutes with interviewers, four times per week.

Transformers, first introduced by~\cite{vaswani2017attention}, leveraging self-attention mechanisms to capture contextual relationships between words in a sequence, have successfully been leveraged in a wide range of NLP tasks including language translation~\cite{camgoz2020sign}, text generation~\cite{brown2020language}, sentiment analysis~\cite{xu2019sentiment}, and more. In addition, Transformers are capable of modeling temporal context and analyzing time series~\cite{zhou2021informer, lim2021temporal, zeng2023transformers}, using self-attention mechanisms to learn contextual relationships across the entire sequence.

\begin{figure*}[t]
  \centering
  \includegraphics[width=1.9\columnwidth]{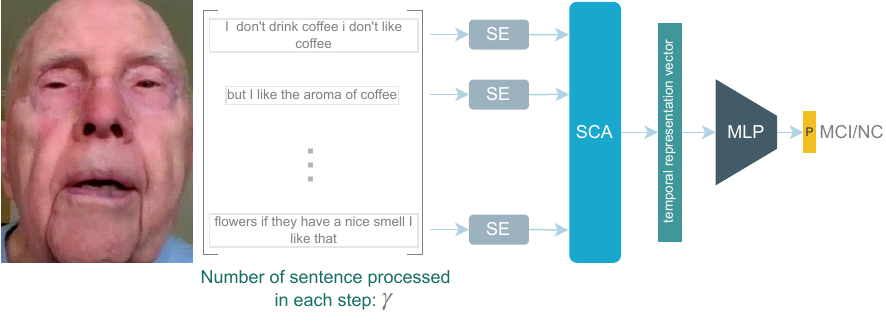}
  \caption{The architecture of our proposed framework. The input to the framework is a sequence of $\gamma$ sentences. The output of the framework is a 2-dimensional vector representing the estimated probabilities of MCI and NC classes. }
  \label{fig:framework_Arch}
\end{figure*}

Our proposed framework consists of two Transformer-based modules, namely \textit{Sentence Embedding} (SE) and \textit{Sentence Cross Attention} (SCA). In the first step, the SE module receives each part of the transcript of each subject's interview, to capture the relation between each word, and finally create a corresponding embedding vector. To clarify further, within each interview, we possess a set of sentences (which can be either complete sentences or phrases). For each of these sentences, we apply the SE module to create the corresponding sentence embeddings. We consider these embeddings as a time series. Thus, we apply the SCA module on these time series to capture the linguistic features for the classification of MCI versus NC. 

While the Cross-Entropy (CE) loss function is widely utilized for classification tasks, to further increase the discriminative power of the framework, we introduce a loss function called \textit{Informative Loss} \textit{(InfoLoss)}, which takes into account the uncertainty associated with ground truth labels during the classification process. Our proposed InfoLoss considers the number of sentences generated by the
participant in each interview as an uncertainty factor. We further show that taking into account these uncertainty factors leads to a more accurate classification (see Sec.~\ref{SEC_Methodology_CustomLoss}).

The contributions of our approach are summarized as follows:
\begin{itemize}

\item We introduce a framework consisting of the SE and SCA modules, designed to capture contextual relationships between words in a sentence, as well as temporal features within a sequence of sentences, to distinguish between MCI and NC.

\item We introduce InfoLoss, a novel loss function that enhances the discriminative power of our framework by taking into account the reduction in entropy by observing each sequence of sentences for each subject.

\end{itemize}

The remainder of this paper is organized as follows. Sec.~\ref{SEC_RelatedWork} reviews the deep learning-based approaches for MCI versus NC classification. Sec.~\ref{SEC_Methodology} describes our proposed NLP-based methodology, the framework architecture, and the proposed loss function for the classification of MCI (vs. NC) based on the transcripts of subjects' interviews. Next, a set of comprehensive experimental results is provided in Sec.~\ref{SEC_Evaluation} to assess the performance of our proposed framework in distinguishing between the MCI and the NC classes. Then, we will discuss how our framework performs the classification task, in Sec.~\ref{SEC_Discussion}. Finally, Sec.~\ref{SEC_Conclusion} concludes the current study and outlines potential future research directions.

\section{Related Work} \label{SEC_RelatedWork}
In this section, we review previous works in detecting neurocognitive impairment using NLP and speech. Utilizing computing power in detecting individuals' cognitive conditions using natural language for accurate and efficient detections of cognitive impairment has a long
research history~\cite{gottschalk1989computerized}. The availability of speech recordings such as the Dementiabank Pitt corpus~\cite{becker1994natural}, and the ADReSS challenge~\cite{luz2020alzheimer} enriched the studies of classifying subjects' cognitive conditions. However, nearly all Deep Learning (DL) models~\cite{fard2023ganalyzer, fard2022sagittal, fard2022acr, fard2022facial, fard2021asmnet, fard2022ad} require large datasets to have an adequate performance over a specific task. Thus, the introduction of Large Language Models (LLM) in the NLP has helped resolve the limited transcripts of AD/ADRD (Alzheimer’s Disease and Related Dementia) patients.

\subsection{Speech-base Detection of AD/ADRD}
The classification of AD/ADRD using speech data has two main approaches: utilizing \textit{acoustic features}, and \textit{linguistic features}. 

\textbf{Acoustic Features:} These features include frequency and spectral aspects of the speech which are dependent on the speaker's vocal cords, and can be extracted using several algorithms~\cite{pappagari2021automatic,pompili2020inesc,bertini2022automatic}. In the following, we review some previously proposed methods, which utilized acoustic features for AD/ADRD detection. Bertini \etal~\cite{bertini2022automatic} proposed an autoencoder to extract features from the Log-Mel spectrogram of subjects in the Pitt corpus. The autoencoder generates latent feature vectors that are classified with an MLP based on the participants' cognitive condition. The utilization of SpecAugment~\cite{park2019specaugment}, a data augmentation method that does not distort the original sample, boosted the model prediction performance.

Acoustic features can effectively capture abnormalities in acoustic features associated with cognitive impairment while remaining applicable regardless of the structure of the spoken language~\cite {yamada2022speech}. However, it should be noted that other neuro diseases can potentially affect acoustic features, and cause an overlap in the acoustic abnormality associated with
cognitive impairment~\cite{fraser2016linguistic}. 

\textbf{Linguistic Features:} These features are extracted from the participants' speech transcript, mostly using LLM (Large Language Models) such as word embedding, syntactic and semantic features, and part of speech. Likewise, advanced language dependency features (i.e. contextual embedding) are illustrated by Deep Learning models such as long short-term memory (LSTM) and Transformers~\cite{rohanian2021multi, roshanzamir2021transformer, ilias2022multimodal}. In the following, we review some previously proposed methods, which utilized linguistic features for AD detection.

Yuan \etal~\cite{yuan2021pauses} have encoded subject pauses during the subject's speech recordings in the ADReSS dataset. The authors investigated and compared the duration of pauses among AD and control group participants. They showed that AD patients tend to have more pauses. They also performed a classification of the subjects' transcripts including and excluding the pauses encoding using Bidirectional Encoder Representations from Transformers (BERT) and Enhanced Representation from kNowledge IntEgration (ERNIE) pre-trained language models (LMs). The models have a maximum input token of 256 to perform a sequence classification. The ensemble of several iterations of the ERNIE model achieved the highest accuracy with the encoding of the pauses within the transcripts data.

Roshanzamir \etal~\cite{roshanzamir2021transformer} used several pre-trained LLMs including BERT, XLNet, XLM, and GloVe word embedding sequence to generate contextual embedding on a sentence level. The logistic regression classifier with BERT$_{LARGE}$ achieved a competitive accuracy on the Pitt corpus dataset in differentiating AD from those with normal cognition. They also integrated the augmenter layer, and bidirectional LSTM during the model fine-tuning but they did not improve the model's performance.

The perplexity score, which measures a language model confidence of generated sample with the probability distribution of the model prediction, is calculated by Colla \etal~\cite{colla2022semantic} to measure the relation between sequence of tokens from a subject with LMs that was fine-tuned by transcripts data of AD and NC excluding the same subject data. The Pitt corpus is used in this study after dropping subjects with only one interview which resulted in 77 AD and 74 HC (healthy control). An LM of each subject is acquired by fine-tuning a pre-trained GPT-2 with a sequence length of 1024 tokens by applying the data from the same class excluding the LM's subject. Also, two LMs are fine-tuned for AD patients and HC participants. The language model generates the probability of a word given its left context which is called the Casual Language Model (CLM). For every subject in the study, two perplexity scores are calculated from both class models. They were able to achieve a perfect performance with 30 epochs of training and using a decision rule of finding a minimum margin between the perplexity score of the predicted class and the different perplexities of the classes' LMs. Eventually, LMs were implemented for every subject and overall LMs of both conditions. 

The linguistic features help the detection of AD patients since humans tend to choose vocabularies and structures of a spoken sentence based on memory. Thus, AD patients have challenges in some aspects of generating a response or describing a picture~\cite{fraser2016linguistic}. Taking that into consideration, it also should be mentioned that linguistic features require computational resources and make the feature interpretation challenging~\cite{colla2022semantic}.

\textbf{Feature Fusion:} There are several methods for merging different features. In the speech signals, fusing is performed after the feature extraction which makes the classifier combine the information. In addition, the fusing can occur after independent models are trained which is referred to as ensemble learning. The feature fusion complements the types of features to make the models have better performance.

Syed \etal~\cite{syed2021automated} combined handcrafted and deep model embedding features for participants' speech and transcript to identify AD patients in the ADReSS dataset. These features are extracted using pre-trained models. Feature aggregation is performed to capture a high-level representation of the data using pooling, bag-of-words, and Fisher Vector Encoding; then, either a Support Vector Machine (SVM) or logistic regression is used for classification. Eventually, a majority voting fusion is performed for a final identification label. The speech transcript has yielded the best classification performance.

Rohanian \etal~\cite{rohanian2021multi} integrated a model that makes decisions from different LSTMs based on the sequential acoustic and linguistic features to detect AD patients for the ADReSS dataset. The acoustic features including prosody, voice quality, and spectral features were extracted using the CONVAREP toolkit~\cite{degottex2014covarep}. Whereas, the linguistic features are extracted as the lexical features from the transcript which GloVe model~\cite{pennington2014glove} is utilized to create word vectors. The best detection performance of the model was achieved by training on both features, the acoustic and linguistic features.

Ilias and Askounis~\cite{ilias2022multimodal} integrated a co-attention mechanism to fuse transcript and image attentions from the ADReSS dataset speech. BERT~\cite{devlin2018bert} and vision transformer (ViT)~\cite{dosovitskiy2020image} are implemented to generate attention among word embeddings and Log-Mel spectrogram images, respectively. A multimodal shifting gate is implemented to integrate multimodal information before predicting the cognitive condition of the ADReSS dataset. The model was able to achieve a high accuracy on the testing set.

Zolnoori \etal~\cite{zolnoori2023adscreen} developed, ADscreen, a speech-based screening tool to detect cognitively impaired patients. The development involves extensive speech enhancement and feature extractions of acoustic and linguistic features from subjects' speech. They studied the following components: acoustic parameters, linguistics parameters, and psycholinguistic cues on the Pitt corpus. The Joint Mutual Information Maximization (JMIM) and Machine Learning models were implemented to select features and classify AD/ADRD, respectively. The acoustic parameters include phonetic motor planning such as MFCC, formant frequencies, and voice intensity. The linguistic features include semantic disfluency in speech, lexical diversity, and the structure of the syntactic. Also, the psycholinguistic cues of patients are identified using Linguistic Inquiry and Word Count (LIWC). The DistilBERT embedded word vectors which helped the ML model join acoustic and linguistic parameters with contextual word embeddings have achieved the best performance with the Support Vector Machine classifier.

\subsection{Review of Research on I-CONECT}
The literature that we examined so far focused on differentiating
those with AD from normal cognition. Several studies have been conducted on the I-CONECT dataset to differentiate the MCI from those with normal cognition. Identification of MCI is more challenging than differentiating between AD and normal cognition due to the subtle decline associated with MCI. The studies focused on the participants' linguistic characteristics~\cite{asgari2017predicting, chen2020topic, tang2022joint} and facial features (such as head movement and facial expression)~\cite{sun2023mc, alsuhaibani2023detection} exclusively. 

Asgari \etal~\cite{asgari2017predicting} categorized participants' spoken words using LIWC into different subcategories of the words such as positive, negative, or filler. They were able to achieve the best performance using the Relatively category with the ML classifier. Chen \etal~\cite{chen2020topic} proposed a method to quantify topical variations based on the lexical coherence of the participants' word selection. They were able to distinguish MCI subjects from those with normal cognition. Tang \etal~\cite{tang2022joint} proposed a method to combine the linguistic (LIWC) and acoustic (MFCC) features of the subjects. They concluded that the detection of the cognitive conditions using the combination of both features outperformed using one feature type exclusively. 

Sun \etal~\cite{sun2023mc} proposed a classifier that boosted the detection performance of the video transformer ViViT~\cite{arnab2021vivit} using the participants' faces. They also proposed a loss function that addresses the inter- and intra-class imbalances of the data. Furthermore, Alsuhaibani \etal~\cite{alsuhaibani2023detection} extracted an interaction feature of participants with interviewers that enhanced the detection of participants' cognitive conditions. They extracted the facial features of subjects on a frame level of the videos by implementing a convolutional autoencoder which generates latent feature vectors. Subsequently, a transformer is implemented to capture the temporal facial features with an integration of the interaction feature.

\section{Methodology} \label{SEC_Methodology}
In this section, we start by presenting a definition of the I-CONECT dataset. Next, we provide an in-depth explanation of the internal modules of our proposed framework, SE, and SCA, followed by introducing the framework. Finally, we present our novel loss function, InfoLoss, and explain how it effectively guides the training process.

\subsection{Definition} \label{SEC_Methodology_Formal}
We define $P^i \in \textbf{P}:\{P^0, P^1, ..., P^N\}$ as the class (MCI or NC) of the $i^{th}$ interviewee person (subject) in the I-CONECT dataset (\textbf{P}). As $P^i$ belongs to either the MCI or NC group, we can define it as follows in Eq.~\ref{eq:s_i_def}:
\begin{align}\label{eq:s_i_def}
\begin{split}
        P^i:= \left\{\begin{matrix} 
                    & 0    & \text{if MCI} \\
                    & 1    & \text{if NC~~}\\
                    \end{matrix}\right. 
\end{split}
\end{align}

For every person $P^i$ in the dataset, we define the transcripts of all conducted interviews belonging to $P^i$, as $R^i \in \textbf{R}:\{R^0, R^1, ..., R^N\}$ ($\textbf{R}$ is the set containing all transcripts in the dataset). As each video in the I-CONECT dataset has a defined theme, which is the topic of the interview (See Sec.~\ref{SEC_Evaluation_I-CONECT}), we can define $R^i$ as follows in Eq.~\ref{eq:r_i_def}:
\begin{align}\label{eq:r_i_def}
\begin{split}
        R^i:= \left\{\begin{matrix} 
                    & theme_0: \{ s_0, s_1, ..., s_l\} \\
                    & theme_1: \{ s_0, s_1, ..., s_q\} \\
                    & \vdots \\
                    & theme_k: \{ s_0, s_1, ..., s_r\} \\
                    \end{matrix}\right. 
\end{split}
\end{align}
Where $l$, $q$, and $r$ are the upper bound for the transcripts associated with $P^i$ and the corresponding themes. As an example, if the number of transcripts in $theme_0$ is 100, then we have $l=100$. We can reduce Eq.~\ref{eq:r_i_def} as follows in Eq.~\ref{eq:r_i_def_reduces}:
\begin{align}\label{eq:r_i_def_reduces}
\begin{split}
        R^i:= \left\{\begin{matrix} 
                    & s_{0,0}, s_{0,1}, ..., s_{0,l} \\
                    & s_{1,0}, s_{1,1}, ..., s_{1,q} \\
                    & \vdots \\
                    & s_{k,0}, s_{k,1}, ..., s_{k,r} \\
                    \end{matrix}\right. 
\end{split}
\end{align}
Accordingly, we introduce $s^{i}_{j,m}$ as the $m^{th}$ sentence in the $j^{th}$ theme corresponding to person $P^i$. The primary objective of the proposed framework is to analyze all sentences associated with an interviewee and categorize the subject as either belonging to the MCI or NC class. 


\subsection{SE Module} \label{SEC_Methodology_SE}
Sentence Transformers~\cite{reimers2019sentence}, are designed to encode a sentence (or a group of words) into a representation vector. The attention mechanism enables the Transformer encoders to process words and their contextual relationships within a sentence and create a representation that captures semantic information, syntactic structure, and contextual nuances~\cite{reimers2019sentence}. 

The SE module in our framework is a pre-trained sentence Transformer. Each input to the SE module is a sentence from a video transcript, and the output is a 768-dimensional representation vector. We use \textit{all-mpnet-base-v2}~\cite{huggingfaceSentencetransformersallmpnetbasev2Hugging}, which is built upon~\cite{song2020mpnet}, as our pre-trained sentence Transformer. This sentence Transformer was trained on very large sentence-level datasets including~\cite{fader2014open, henderson2019repository, lo-etal-2020-s2orc} using a self-supervised contrastive learning objective. The primary goal of the SE module is to create a \textit{sentential representation vector} for each sentence that encapsulates the semantic information, to be utilized in the downstream task by the SCA module. Likewise, we concatenate the speech duration of each input phrase with the output sentence embedding, to create sentential representation vectors. Finally, we introduce $E^{i}_{j,m}$ the $m^{th}$ sentential representation vector, in the $j^{th}$ theme belonging to $R^i$.

\subsection{SCA Module} \label{SEC_Methodology_SCA}
\begin{figure}[t]
  \centering
  \includegraphics[width=0.7\columnwidth]{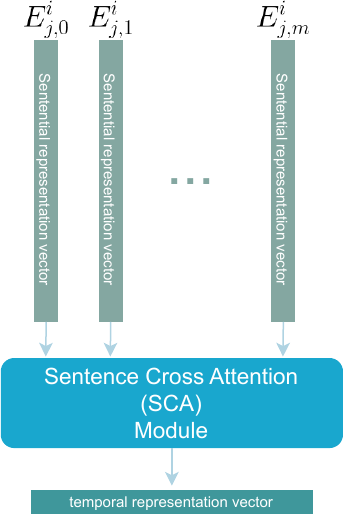}
  \caption{This figure shows the SCA module. The input to the SCA module is a sequence of sentential representation vectors, and the output is the corresponding temporal representation vector.}
  \label{fig:sca_module}
\end{figure}

Sentential representation vectors are time series that contain semantic information, and contextual nuances associated with each video transcript. We designed the SCA module for analyzing these time series, capturing intricate temporal relationships and dependencies, and ultimately modeling patterns that can be used for classifying MCI versus NC. The SCA module is a Transformer encoder, where its input is a sequence of $\gamma=200$ number of sentential representation vectors, and its output is called \textit{temporal representation vector}, which eventually represents the features that can be used in the downstream classification task. Fig.~\ref{fig:sca_module} represents the SCA module. 

The SCA module is a 1-layer Transformer encoder~\cite{vaswani2017attention}, except there is no positional embedding in the architecture. Particularly, Transformers usually employ positional encoding during word processing to differentiate between identical words occurring multiple times within a phrase, while the inputs to the SCA module are identical sentential representation vectors. Hence, as the inputs to this Transformer are sentential representation vectors that are generated by the SE module, there is no need for positional encoding.  The output size of the Transformer encoder is the same as its input (\textit{(1+size of sentence embedding)}~$\times~\gamma$). Thus, a pooling layer is used to create the final temporal representation vector, which is a 769-dimensional vector. Please refer to Sec.~\ref{SEC_Evaluation_implementation} for details about the architecture, implementation, and training of the SCA module.

\subsection{Framework Architecture} \label{SEC_Methodology_Framework_Arch}
As Fig.~\ref{fig:framework_Arch} shows, our proposed architecture consists of $\gamma$ number of SE modules, one SCA module, and an MLP. For each $P^i$, we feed the architecture a sequence of $\gamma$ sentences and train both the SCA module and MLP simultaneously to predict $\hat{P}_i$, the corresponding class label. The MLP consists of 2 \textit{Dense} layers, with sizes of 384, and 2, respectively, followed by a Softmax activation function. In other words, we designed the proposed framework to analyze a sequence of sentences, which is the transcriptions of videos conducted during each interview in the I-CONECT project, with the aim to classify the corresponding object into either the MCI or NC group. Please refer to Sec.~\ref{SEC_Evaluation_implementation} for more details about the model training.

\subsection{InfoLoss} \label{SEC_Methodology_CustomLoss}
In this section, we first define the objective function and the optimization of the model. Then, we introduce the Frequency-Based Uncertainty to be utilized in our framework. Finally, we propose our novel loss function, leveraging the discussed uncertainty in the framework for a more accurate classification of the MCI versus NC.

\subsubsection{Objective Function and Inference Classification} \label{SEC_Methodology_CustomLoss_objective}
As explained in Sec.~\ref{SEC_Methodology_Formal}, for each person $P^i$ in the training set, there exists $R^i$, as the set of transcripts of videos regarding that person. As Fig.~\ref{fig:framework_Arch} shows, our framework is designed to analyze a sequence of $\gamma$ sentences that are part of $R^i$ and predict the class label $\hat{P}^{i}_{j, m}$, where $j$, and $m$ refer to the theme and the starting index of the sentence in the sequence, respectively. For simplicity, we introduce $seq^{i}_{j, m}$ as the sequence of sentences indexed from $m$ to $m+\gamma$ of the $j^{th}$ theme in $R^i$ as follows in Eq.~\ref{eq:sequence}:
\begin{equation}\label{eq:sequence}
 seq^{i}_{j, m} := ~\{s^{i}_{j, m}, ..., s^{i}_{j, m+\gamma}\}
\end{equation}
Accordingly, the objective function of our proposed framework is to minimize $|P^i - \hat{P}^{i}_{j, m}|$, given an arbitrary sequence of sentences $seq^{i}_{j, m}$ from $R^i$.

Our proposed framework is theme-agnostic, which means we do not consider the themes as a condition when training the framework. Hence, we can simply define $Seq^{i}$ as the set of all sequences of sentences regardless of the theme, associated with the $i^{th}$ subject, $P^i$. In inference, we introduce the predicted class label $\hat{P}^{i}$ (using the proposed framework) corresponding to person $P^i$, as the average of the predicted probabilities of all the sequences in $Seq^{i}$ (all the sequences within $R^i$, regardless of theme) for each class label (MCI versus NC), as follows in Eq.~\ref{eq:p_prime}:
\begin{align}
& \begin{adjustbox}{width=190pt}$
\hat{P}^{i}_{MCI} := \sum_{}^{} \mathcal{F}( MCI~|~seq) ~~~~~ \forall seq \in Seq^{i}
$ \end{adjustbox}   \\
&\begin{adjustbox}{width=190pt}$
\hat{P}^{i}_{NC} :=  \sum_{}^{} \mathcal{F}( NC~|~seq) ~~~~~ \forall seq \in Seq^{i}
$ \end{adjustbox} 
\end{align}
\begin{align}\label{eq:p_prime}
\begin{split} 
        \hat{P}^{i} := \left\{\begin{matrix} 
                        0 &\text{if} & \hat{P}^{i}_{MCI} > \hat{P}^{i}_{NC} \\
                        1 &\text{otherwise} 
                         \end{matrix}\right. \\
\end{split}
\end{align}
where $\mathcal{F}$ is our proposed framework. As it can be interpreted from Eq.~\ref{eq:p_prime}, the final predicted class label $\hat{P}^{i}$, heavily relies on the sequence-level probability scores ($\hat{P}^{i}_{j, m}$). 

It is noteworthy to highlight that most previous studies~\cite{sun2023mc, alsuhaibani2023detection} on I-CONECT which proposed a sequence-based framework, employed a majority voting strategy during inference to determine the class label based on available sequences. In our methodology, we consider each prediction for every sequence of sentences as a sole probability score. Consequently, we calculate the average probability scores across all sequences, as shown in~Eq.~\ref{eq:p_prime}, for each subject to set the ultimate associated class label. While majority voting ignores the confidence of the model (the probability score assigned to each input sequence by the model) for each prediction, our approach takes into account the confidence of each prediction, providing a more robust prediction.

\begin{figure*}[t]
  \centering
  \includegraphics[width=2\columnwidth]{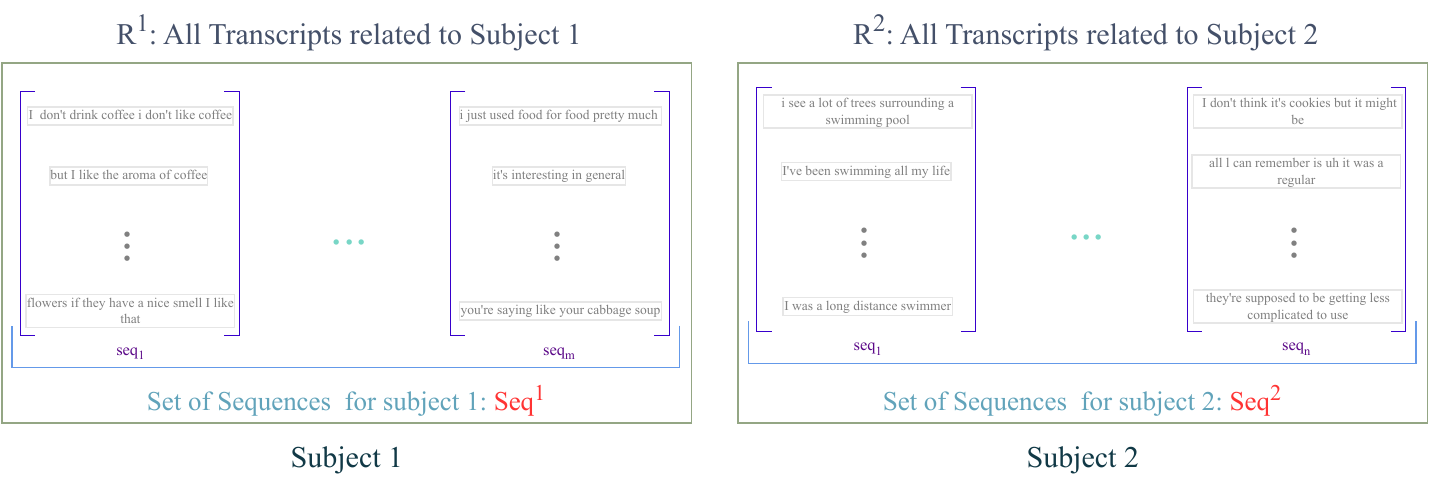}
  \caption{The figure shows an example of sequences of sentences for subjects in the I-CONECT dataset. Assuming \textit{m} is the number of sequences for subject 1, and \textit{n} is the number of sequences for subject 2. If \textit{m} is smaller than \textit{n}, observing the class label for each sequence related to Subject 1 provides more information about the ultimate class label for Subject 1 than observing each sequence related to Subject 2. In simpler terms, observing each sequence associated with Subject 1 contributes more to reducing the entropy of Subject 1 compared to the entropy reduction of Subject 2 performed by observing its related sequences. }
  \label{fig:uncertainty}
\end{figure*}

\subsubsection{Frequency-Based Uncertainty} \label{SEC_Methodology_CustomLoss_Freq}
As explained in Sec.~\ref{SEC_Methodology_CustomLoss_objective}, the input to our framework is a sequence of $\gamma$ sentences, and the framework predicts the probability score associated with that sequence regarding each class. Subsequently, the ultimate probability score for an arbitrary subject $P^i$ is the average probability score of all the sequences $Seq^{i}$ within $R^i$.

From the Information Theory perspective, we have a set of evidence (set $Seq^{i}$, which contains $k$ sequence of sentences, associated with person $P^i$), and by observing all these sequences, we decide to label the corresponding subject (a person). In other words, observing each sequence of sentences belonging to a subject, reduces the entropy regarding the class label of that subject, thereby, we can more confidently predict the class of the subject. Consequently, the more we observe sequences, the more information regarding the corresponding subject is gained, and the more confidently a subject is classified to be either MCI or NC. Please refer to \nameref{SEC_Appendix} for more clarification.

From another perspective, we can infer that the quantity of sequences of sentences available for a particular subject exhibits a strong correlation with the information that each sequence can unveil, which is the same as the amount of reduction in entropy associated with the class of a subject (Please refer to \nameref{SEC_Appendix} for the corresponding proof). Therefore, the significance of observing each sequence in reducing entropy regarding the class of a subject $P^i$ is inversely correlated to the occurrence rate of those sequences, which is the count of sequences within associated the set of transcripts, $R^i$. In other words, in subjects with fewer sequences, observing each sequence results in a higher reduction in the corresponding entropy, compared to the subject with more sequences. Fig.~\ref{fig:uncertainty} illustrates how the occurrence frequency of sequences related to a subject plays a role in reducing the uncertainty (entropy) associated with the classification of that specific subject. In simpler terms, it demonstrates how the number of sequences in transcripts regarding a subject, impacts the amount of information that observing each sequence can provide for ultimate classification between MCI and the NC group.

We claim that since each sequence of sentences cannot provide enough information about the class of subject it belonged to, we need to train the framework considering this uncertainty. More clearly, instead of using the original binary ground truth labels (0 for MCI and 1 NC), we propose a novel approach that introduces label smoothing by taking into account the amount of information (thereby, reduction in entropy), that each sequence of sentences can provide for classification of each subject.

For each class, we can model the distribution of the corresponding sequences with a multivariate Bernoulli distribution. To clarify further, $p^{i}_{j,m}$, the ground truth label for an arbitrary sequence associated with subject $i$ is a 2-dimensional vector, where each index can be either zero or one depending on the subject label as follows in Eq.~\ref{eq:seq_bernoulli}:
\begin{align}\label{eq:seq_bernoulli}
\begin{split} 
        p^{i}_{j,m} := \left\{\begin{matrix} 
                        \begin{bmatrix} 1\\ 0\end{bmatrix} &\text{if} & P^{i} \text{~~is MCI} \\ \\
                        \begin{bmatrix} 0\\ 1\end{bmatrix}  &\text{if} & P^{i} \text{~~is NC}
                         \end{matrix}\right. \\
\end{split}
\end{align}
With this definition, we implicitly assume that the observation of each sequence provides us with sufficient information to label its corresponding subject (\textit{i.e.} observing one sequence of $\gamma$ sentences gives us adequate information to classify a subject as MCI, or NC). However, for a more reliable classification of an arbitrary subject, our framework takes into consideration all the corresponding sequences. 

In our approach, we introduce a modification to the original Bernoulli distribution of ground truth labels during the training phase. This modification is aimed at introducing a level of smoothness to the distribution. To implement this, we introduce a hyper-parameter, denoted as $\varepsilon$, with a value of $0.2$ acting as the smoothing factor. This hyper-parameter plays a critical role in transforming the Bernoulli distribution (refer to Eq.~\ref{eq:seq_bernoulli}) into a smoothed distribution. To be more detailed, for any sequence associated with a subject classified as MCI, we revise the Bernoulli label defined in Eq.~\ref{eq:seq_bernoulli}) as $[1-\varepsilon, \varepsilon]^{T}$. This basically means that we soften the ground truth label for the MCI class using ($\varepsilon$). Similarly, for sequences belonging to subjects that are labeled as NC, we revise the Bernoulli defined in Eq.~\ref{eq:seq_bernoulli} as $[\varepsilon, 1-\varepsilon]^{T}$. It is important to mention that this modification only occurs during the training process, and only for training models. Thus, we can revise Eq.~\ref{eq:seq_bernoulli}, and introduce Eq.~\ref{eq:seq_bernoulli_smooth_dist} as follows:
\begin{align}\label{eq:seq_bernoulli_smooth_dist}
\begin{split} 
        p^{i}_{j,m} := \left\{\begin{matrix} 
                        \begin{bmatrix} 1 - \varepsilon\\  \varepsilon\end{bmatrix} &\text{if} & P^{i} \text{~~is MCI} \\ \\
                        \begin{bmatrix} \varepsilon\\ 1- \varepsilon\end{bmatrix}  &\text{if} & P^{i} \text{~~is NC}
                         \end{matrix}\right. \\
\end{split}
\end{align}
Although this revised definition of $p^{i}_{j,m}$ reflects that an arbitrary sequence of sentences cannot present enough information to be used exclusively for reliable classification of its corresponding subject, it still ignores the amount of information each sequence can provide. For clarification, assuming the length of $Seq^i$ associated with person $P^i$ is $l^i$, and the length of $Seq^j$ associated with person $P^j$ is $l^j$. As discussed, if $l^i$ is greater than $l^j$, then the sequences of sentences belonging to $Seq^j$ contribute more to the reduction of entropy regarding $P^j$ compared to the sequences of sentences belonging to $Seq^i$. Hence, we need to define $p^{i}_{j,m}$ for each subject in the training set, considering the amount of information --which is equivalent to the reduction in the entropy-- that observing each sequence of sentences can provide. 

For an arbitrary subject $P^i$, we use the inverse relationship between the frequency of sequences within $R^i$ (which is equal to $|Seq^i|$) and the amount of information that observing each sequence provides, to define an uncertainty factor $\psi^i$. Likewise, for every ground truth subject in $\textbf{P}:\{P^0, P^1, ..., P^N\}$, we compute the corresponding uncertainty factors $\Psi = \{ \psi^0,...,  \psi^N\}$. Finally, introduce the smooth ground truth $P_{smooth}^i$, associated with the $i^{th}$ subject as follows in Eq.~\ref{eq:seq_bernoulli_soft}:
\begin{align}\label{eq:seq_bernoulli_soft}
\begin{split} 
        P_{smooth}^i := \left\{\begin{matrix} 
                        \begin{bmatrix} 1 - \psi^i\\ \psi^i\end{bmatrix} &\text{if} & P^{i} \text{~~is MCI} \\ \\
                        \begin{bmatrix} \psi^i\\ 1-\psi^i\end{bmatrix}  &\text{if} & P^{i} \text{~~is NC}
                         \end{matrix}\right. \\
\end{split}
\end{align}

To calculate $\psi^i$, the uncertainty factor associated with person $P^i$, we introduce the \textit{frequency set} called $\mathbf{SF}:= \{ |Seq^{0}|, |Seq^{1}|, ..., |Seq^{N}|\}$. Basically, the $i^{th}$ element of $\mathbf{SF}$ represents the length of $Seq^{i}$ associated with the person $P^{i}$ in the training set. Finally, by simply map $\mathbf{SF}$ to be in the range $[\varepsilon-\frac{\varepsilon}{2}, \varepsilon+\frac{\varepsilon}{2}]$ (that is [0.1, 0.3]), we calculate $\psi^i$ as follows in Eq.~\ref{eq:psi_calculation}:
\begin{align}\label{eq:psi_calculation}
\psi^i := \frac{\varepsilon(|Seq^{i}|- \frac{\varepsilon}{2})}{ Max(\mathbf{SF}) -  Min(\mathbf{SF})} +  \frac{\varepsilon}{2} 
\end{align}
Therefore, the uncertainty factor $\psi^a$ for person $P^a$ is smaller than $\psi^b$ for person $P^b$ if there are fewer transcripts available for $P^a$ compared to $P^b$ in the training set.


The process involves calculating $\psi^i$ as defined in Eq.\ref{eq:psi_calculation} and then applying Eq.\ref{eq:seq_bernoulli_soft} to generate a smooth ground truth label corresponding to person $P^i$. This ensures that the training process takes into account the uncertainty associated with the sequence of sentences. More clearly, instead of expecting each sequence of sentences to provide enough information to classify a person, by training the model with smooth ground truth labels, we consider the intrinsic uncertainty of the problem. In Sec.~\ref{SEC_Methodology_CustomLoss_defin}, we define the loss function using our proposed frequency-based uncertainty.

\subsubsection{Loss Function} \label{SEC_Methodology_CustomLoss_defin}
As discussed in Sec.\ref{SEC_Methodology_CustomLoss_Freq}, we smooth the labels for each sequence by utilizing our proposed frequency-based uncertainty. Thus, instead of having a static class label for a sequence, we have a distribution. Hence, we train our framework to generate distributions that closely resemble the ground-truth distributions associated with each sequence. To minimize the difference between an arbitrary ground-truth distribution, and the corresponding predicted distribution, we use the Kullback–Leibler divergence (KLD)~\cite{kullback1951information}.

We show in Sec.~\ref{SEC_Evaluation_Ablation} that utilizing the frequency-based uncertainty, and thereby, applying KLD instead of CE results in a more accurate classification.

\section{Evaluation and Experimental Results} \label{SEC_Evaluation}
In this section, we first give a detailed overview of the I-CONECT dataset. Then, we illustrate the implementation details of the framework and the training, followed by an introduction to the evaluation metrics. Finally, we provide a comprehensive experimental result and the ablation study.

\begin{table}[b!] 
\caption{ Distribution of automatic and manual transcripts in the I-CONECT dataset, including counts and proportions.}
\label{tbl:tbl_tr_type}
\centering
\small
\begin{tabular}{lcc}
\hline
Transcripts Type                              & Number & Portion \\ \hline
Automatic Transcripts & 6272   & 99.55\% \\
Manual Transcripts    & 28     & 0.45\%  \\ \hline
\end{tabular}
\end{table}

\subsection{I-CONECT Dataset} \label{SEC_Evaluation_I-CONECT}
I-CONECT aimed to investigate the potential of social dialogue in enhancing memory and potentially mitigating the onset of dementia or Alzheimer's disease among older adults. This research initiative followed individuals aged 75 and above, selected from Portland, Oregon, or Detroit, Michigan in the USA. The study involved the random allocation of 187 participants into two groups, control and experimental. The control group received weekly telephone check-up calls whereas the experimental group participated in semi-structured conversations via video chat with standardized interviewers on pre-selected topics. The participants are distinguished by their cognitive conditions of MCI or NC. During six months, the participants engaged in 30-minute conversations four times with standardized interviewers per week. All participants utilized user-friendly devices provided by the study to communicate with the conversational staff. 

Initiated with standard prompts and daily topics, these interactions involved semi-structured conversations that allowed for enjoyable and natural conversational exchanges. Recorded in the form of videos, these conversational interactions encompassed a range of 161 selected themes, including Health Care, Summertime, Military Service, etc., within each half-hour session. Among the 68 randomized individuals, 34 were identified with MCI, while 34 were diagnosed with NC conditions. 

The I-CONECT study has been partially manually transcribed from the video recordings. The remaining of the study videos are transcribed using Automated Speech Recognition (ASR). Chen and Asgari~\cite{chen2021refining} have refined an ASR system to improve the automated transcript of older adults using the manual transcript data. Thus, we have access to the automated and manual transcript of the data. It is worth mentioning that our training and testing separation do not take into consideration the original method of generating the transcripts. Table~\ref{tbl:tbl_tr_type} provides details of automatic and manual transcripts in the I-CONECT dataset.

\begin{figure}[t]
  \centering
  \includegraphics[width=0.95\columnwidth]{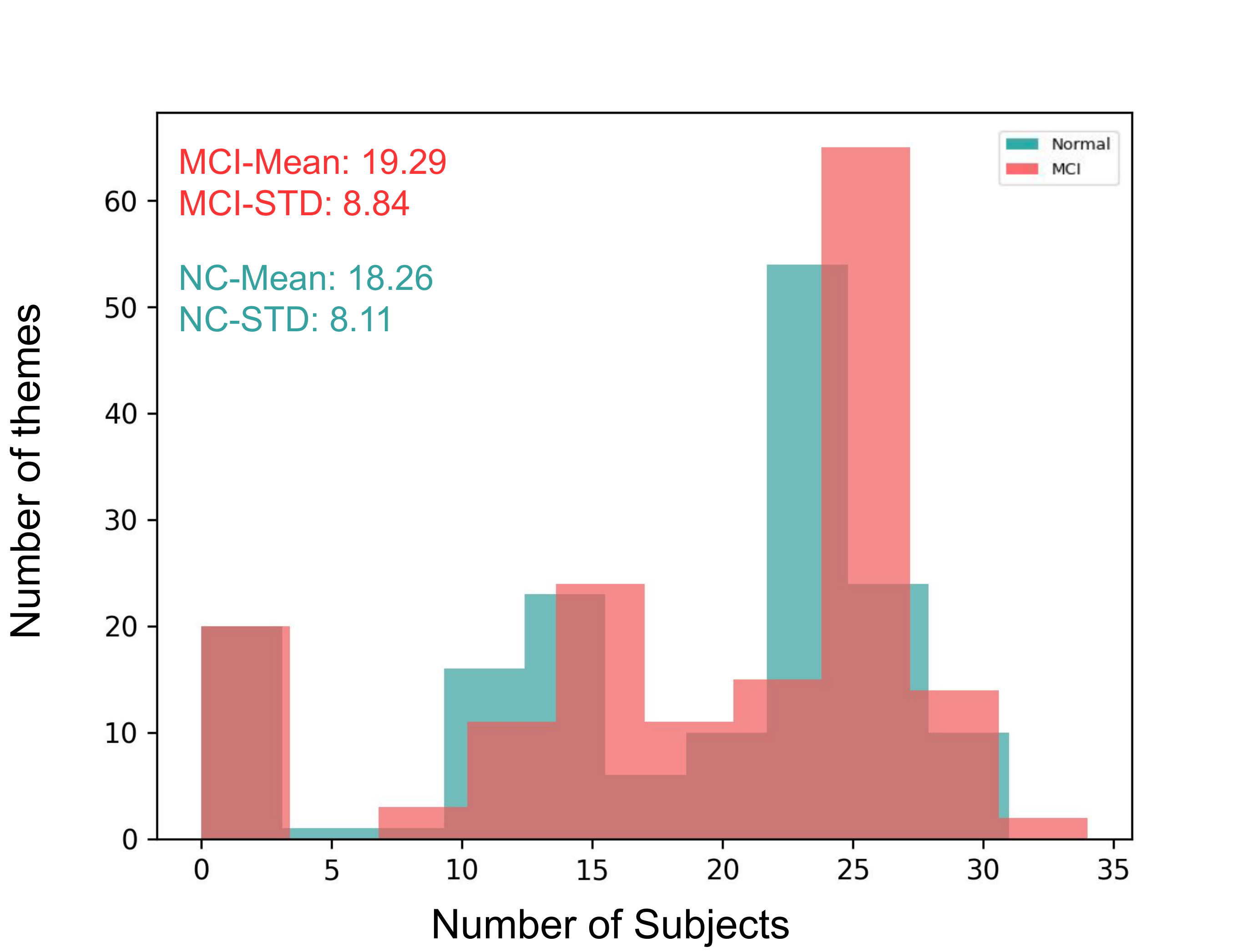}
  \caption{The distribution of themes per subject among individuals with MCI and NC.}
  \label{fig:ana_Theme_subject_hist}
\end{figure}

\begin{table}[b] 
\caption{ The configuration of the folds, including the number of subjects in the training set and test set.}
\label{tbl:tbl_fold_conf}
\centering
\small
\begin{tabular}{cc|ccccc}
\hline
\multicolumn{2}{c|}{Fold\#}            & 1 & 2 & 3 & 4 & 5 \\ \hline
\multirow{3}{*}{\rotatebox{0}{Train}}
            & MCI & 29  & 28  & 28  & 28  & 28  \\
            & NC  & 27  & 27  & 27  & 27  & 28  \\
            & sum & 56  & 55  & 55  & 55  & 56  \\ \hdashline
\multirow{3}{*}{\rotatebox{0}{Test}}
            & MCI &6   &7   & 7   & 7   & 7   \\
            & NC  &7   &7   & 7   & 7   & 6  \\
            & sum &13  &14  &14   &14   & 13  \\ \hline
\end{tabular}
\end{table}

\begin{figure}[t]
  \centering
  \includegraphics[width=1.0\columnwidth]{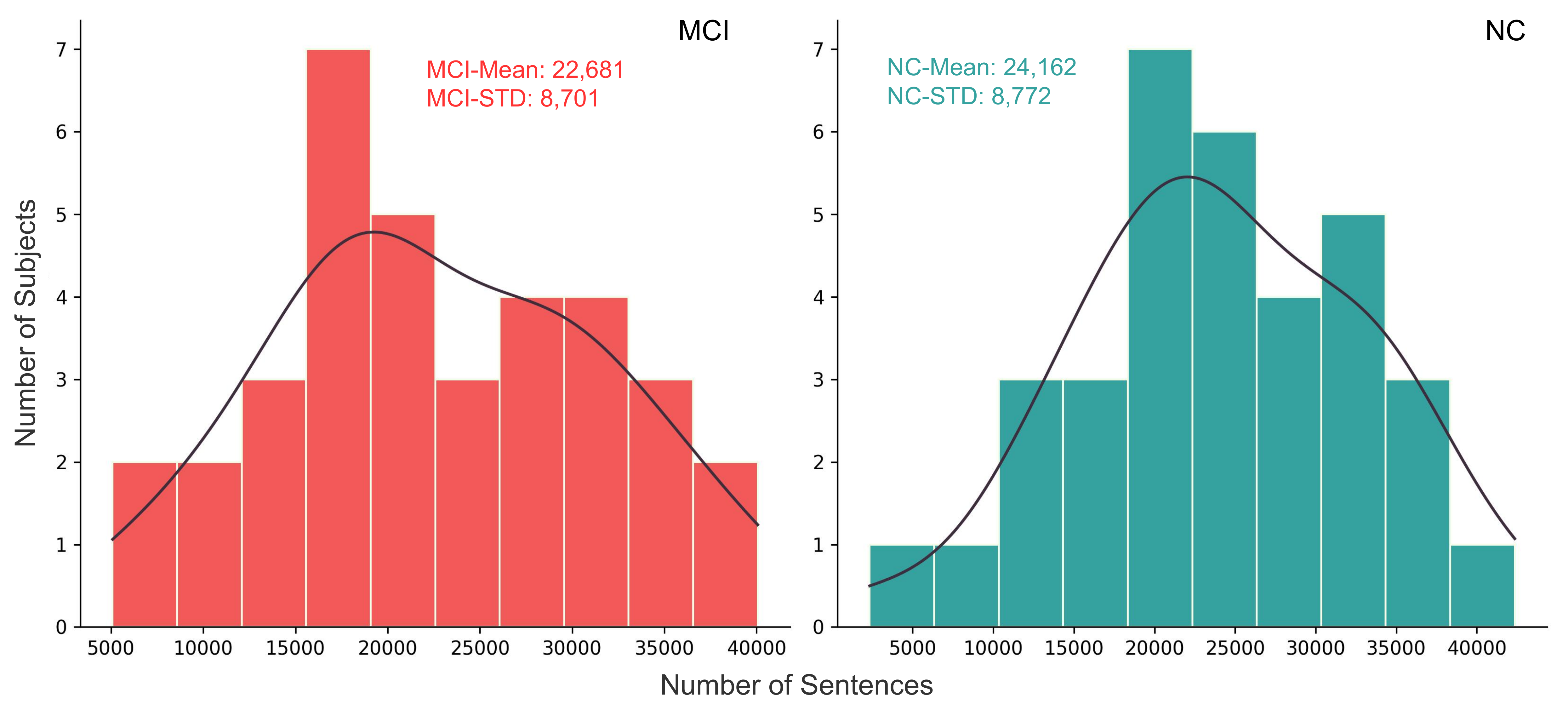}
  \caption{The distribution of sentences per subject for MCI (left chart) and NC (right chart) groups.}
  \label{fig:ana_sen_sub_hist}
\end{figure}

To begin, we present an analysis based on themes and subjects, outlining the distribution of the subjects participating in each theme in Fig.~\ref{fig:ana_Theme_subject_hist}. Accordingly, in general, the MCI class has slightly more participation compared to the NC group. More specifically, as Fig.~\ref{fig:ana_Theme_subject_hist} presents, the average number of themes each MCI subject participated in is 19.29, while this number is 18.26 for the NC group. Since the number of sentences within each theme might defer, we provide a sentence-level analysis as well. Hence, in Fig.~\ref{fig:ana_sen_sub_hist}, we present the histogram illustrating the distribution of the number of sentences for each subject. According to Fig.~\ref{fig:ana_sen_sub_hist}, regardless of theme, the average number of sentences for each subject in the MCI group is about 22.68, with a standard deviation (std) of around 8.7, while these numbers are 24.16, and 8.77 for the NC class respectively.

\subsection{Implementation Detail} \label{SEC_Evaluation_implementation}
\textbf{The SE Module:} As explained in Sec.~\ref{SEC_Methodology_SE}, the SE module is a pre-trained Transformer designed to create a sentence embedding from a set of words. The input to this module can be up to x number of words and the generated sentence embedding is a vector of size 768. We concatenate the speech duration regarding each transcript as the first element of the sentence embedding, thereby, the output vector size is 769.

\textbf{The SCA Module:} This module is a 1-layer Transformer encoder, without the positional embedding. The SCA Transformer has 8 heads, and the size of the fully connected layer used in this module is 128, with the dropout rate set to 0.3. In the last layer, we utilized a global average pooling layer, and hence, the output is a 769-dimensional vector. We apply the Sigmoid activation function to confine each element of the output vector to be in the [0,1] range. 

\textbf{The MLP:} We structure the MLP with 2 fully connected layers, having dimensions of 384, and 2, respectively. The first layer is activated by the ReLU activation function. The last layer is utilized with the Softmax activation function to generate the class label probabilities.

\textbf{Training:} For the training process, we conduct approximately 100 epochs for both the SCA and MLP modules, employing two separate Adam optimizers~\cite{kingma2014adam} with the learning rate set at $10^{-4}$, $\beta_1=0.9$, $\beta_2=0.999$, and decay=$10^{-7}$. Additionally, we configure the warm-up step to be 0.1 of the total training epochs. We implemented our framework using the TensorFlow library and ran it on an NVidia 1080Ti GPU. 

\subsection{Evaluation Metrics} \label{SEC_Evaluation_val_metrics}
We employ a comprehensive set of evaluation metrics in assessing the performance of our proposed framework in distinguishing between MCI and NC. Firstly, we report \textit{accuracy} (Acc), which measures the overall proportion of correctly classified instances. For a balanced assessment of model performance, the F-1 score, which combines precision and recall, is also reported. We also report the Area Under the Receiver Operating Characteristic Curve (AUC), which quantifies the framework's ability to distinguish between the two classes across different thresholds. Furthermore, we provide the confusion matrix, which represents the true positive, true negative, false positive, and false negative explicitly.

\subsection{Experiments} \label{SEC_Evaluation_Experiments}
Since the number of available subjects in the I-CONECT dataset is 68, we assess the performance of our framework using the K-fold validation technique. We define 5 folds, and thereby, for each fold, we allocate 20\% of the subjects randomly as the test set, and the remaining subjects for the training set. To ensure fair evaluation, we designed the test set to be balanced in most folds. Specifically, half of the subjects in the test set are randomly chosen from the MCI group, and half from the NC, resulting in a relatively balanced test set. Table.~\ref{tbl:tbl_fold_conf} shows the configuration of the folds for the training and test set. In addition, we assess the performance of the proposed framework for each fold while providing the average performance of all folds.

\textbf{Accuracy and AUC:} Table~\ref{tbl:tbl_acc_auc} presents the accuracy and the AUC values on a per-fold basis, along with the average of these metrics. As the test set tends to be balanced in most cases, the values for accuracy and AUC are very close. According to Table~\ref{tbl:tbl_acc_auc}, the framework exhibited its lowest performance on fold 1, with accuracy and AUC values of 76.92\% and 76.19\%, respectively, whereas the highest performance is achieved on fold 2, where both accuracy and AUC values reached 92.85\%. For folds 3, 4, and 5, both accuracy and AUC values are close to 85\%. Consequently, as indicated in Table~\ref{tbl:tbl_acc_auc}, the framework achieves an overall accuracy of 85.16\% and an AUC of 84.75\%, with the corresponding std of 5.76 and 5.97, respectively. 
\begin{table}[t] 
\caption{Subject-level accuracy, and AUC values of the proposed framework for each fold. The average values and the standard deviation of all 5 folds are also provided.}
\label{tbl:tbl_acc_auc}
\centering
\small
\resizebox{0.45\textwidth}{!}
{{
\begin{tabular}{c|ccccc|cc}
\hline
Fold\# & 1 & 2 & 3 & 4 & 5 & Avg & std\\ \hline
Acc(\%)  & 76.92  & 92.85  & 85.71  & 85.71   & 84.61  & 85.16 & 5.76 \\
AUC(\%)  & 76.19  & 92.85  & 85.71  & 85.71   & 83.33  & 84.75 & 5.97 \\ \hline
\end{tabular}
}}
\end{table}
\begin{table}[t] 
\caption{Subject-level precision, recall, and F-1 score values of the proposed framework for each fold. The average values and the standard deviation of all 5 folds are also provided.}
\label{tbl:tbl_f1_pre_rec}
\centering
\small
\resizebox{0.45\textwidth}{!}
{{
\begin{tabular}{cc|ccccc|cc}
\hline
\multicolumn{2}{c|}{Fold\#}      & 1     & 2     & 3     & 4     & 5     & \multicolumn{1}{c}{Avg} & \multicolumn{1}{c}{std} \\ \hline                           
\multirow{2}{*}{Precision} 
& MCI & 80    & 100   & 85.71 & 100   & 77.78       & 88.69  & 10.71                   \\
& NC  & 75    & 87.50 & 85.71 & 77.77 & 100         & 85.19  & 9.79                   \\ \hdashline
                           
\multirow{2}{*}{Recall}    
& MCI & 66.66 & 85.71 & 85.71 & 71.42 & 100         & 81.09  & 13.21                   \\
& NC  & 85.71 & 100   & 85.71 & 100   & 66.66       & 87.61  & 13.72                   \\ \hdashline

\multirow{2}{*}{F-1}       
& MCI & 72.72 & 92.30 & 85.71 & 83.33 & 87.50       & 84.31   & 7.26                    \\
& NC  & 80    & 93.33 & 85.71 & 87.50 & 80.00       & 85.30   & 5.60                   \\ 
                           \hline
\end{tabular}
}}
\end{table}

\textbf{Precision, Recall, and F-1 score:} In Table~\ref{tbl:tbl_f1_pre_rec}, we provide a comprehensive representation of per-class and per-fold metrics, including precision, recall, and F-1 scores. Concerning precision, our framework achieved 88.69\% for the MCI class, and 85.19\% for the NC class, representing the ability of the framework to make accurate positive predictions for both classes, with slightly higher (about 3.5\%) accuracy for the MCI class. For the MCI class, the average value of recall is 81.09\%, while the average recall for the NC class is 87.61\%, indicating the performance of the framework at minimizing false negatives rate. Likewise, the Sensitivity of our framework is slightly higher (about 6.52\%) for the NC class. Regarding the F-1 score, our proposed framework achieves an average performance of 84.31\% for the MCI class and 85.30\% for the NC class, indicating the effectiveness of the framework in distinguishing between both classes. The negligible difference (about 0.99\%) between per-class F-1 scores highlights the effectiveness of our framework in achieving a balanced measure of precision and recall for both classes.

\begin{table}[t] 
\caption{Sequence-level accuracy, and AUC values of our proposed framework for each fold. The average values and the standard deviation of all 5 folds are also provided.}
\label{tbl:tbl_acc_auc_seq}
\centering
\small
\resizebox{0.45\textwidth}{!}
{{
\begin{tabular}{c|ccccc|cc}
\hline
Fold\# & 1 & 2 & 3 & 4 & 5 & Avg & std\\ \hline
Acc(\%)  & 72.33  & 79.46  & 69.05  & 71.09   & 70.35       & 72.45 & 4.09 \\
AUC(\%)  & 72.31  & 79.14  & 68.90  & 71.55   & 70.06        & 72.39 & 3.99 \\ \hline
\end{tabular}
}}
\end{table}

\begin{table}[t] 
\caption{Sequence-level precision, recall, and F-1 score values of the proposed framework for each fold. The average values and the standard deviation of all 5 folds are also provided.}
\label{tbl:tbl_f1_pre_rec_seq}
\centering
\small
\resizebox{0.45\textwidth}{!}
{{
\begin{tabular}{cc|ccccc|cc}
\hline
\multicolumn{2}{c|}{Fold\#}      & 1     & 2     & 3     & 4     & 5     & \multicolumn{1}{c}{Avg} & \multicolumn{1}{c}{std} \\ \hline                           
\multirow{2}{*}{Precision} 

& MCI & 74.84    & 72.60   & 75.54   & 64.44  & 65.70       & 70.62  & 5.20     \\
& NC  & 70.21    & 85.10   & 61.72   & 77.06  & 74.87       & 73.79  & 8.63     \\ \hdashline
                           
\multirow{2}{*}{Recall}    
& MCI & 67.51   & 81.64   & 68.10    & 68.55  & 74.25       & 72.01  & 6.02     \\
& NC  & 77.14   & 77.29   & 70.00    & 73.64  & 66.45       & 72.90  & 4.68     \\ \hdashline

\multirow{2}{*}{F-1}       
& MCI & 70.99   & 76.86   & 71.63    & 66.43  & 69.71       & 71.12   & 3.78    \\
& NC  & 70.21   & 81.10   & 65.60    & 75.31  & 70.41       & 72.52   & 5.89    \\ \hline

\end{tabular}
}}
\end{table}
\begin{figure*}[t!]
  \centering
  \includegraphics[width=1.99\columnwidth]{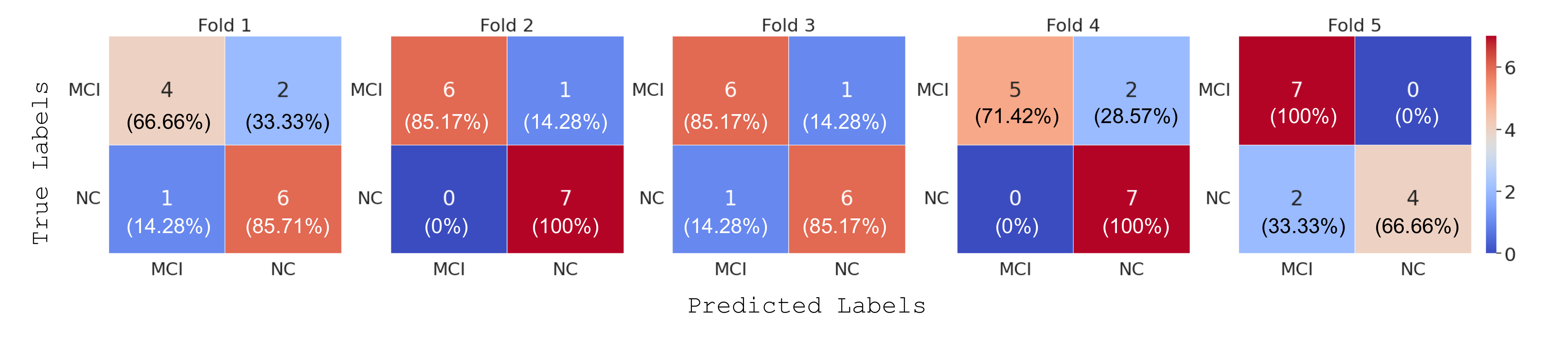}
  \caption{Subject-level confusion matrices for each fold.}
  \label{fig:conf_matrix}
\end{figure*}

\textbf{Confusion Matrices:} Fig.~\ref{fig:conf_matrix} shows the confusion matrices for each fold. As depicted in Fig.~\ref{fig:conf_matrix}, our proposed framework shows an overall superior performance in accurately classifying the NC class compared to MCI. Specifically, for folds 1, 2, and 4, the accuracy values of the classification for the NC class are 85.71\%, 100\%, and 100\% respectively, while the corresponding values for the MCI class are 66.66\%, 85.17\%, and 71.42\% respectively. Regarding fold 3, the framework performs equally for both classes. Only for fold 5, the framework performs better in correctly classifying the MCI class, with an accuracy value of 100\%, compared to the NC class, with an accuracy value of 66.66\%.

\begin{figure*}[t]
  \centering
  \includegraphics[width=1.99\columnwidth]{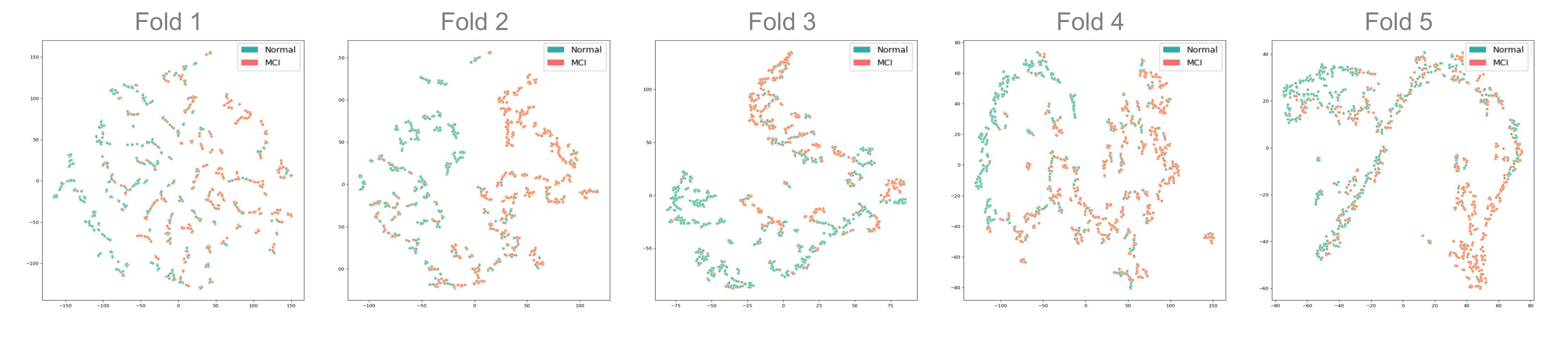}
  \caption{The temporal representation vectors regarding MCI and NC classes for each fold.}
  \label{fig:embedding_visualization}
\end{figure*}

\textbf{Visualization of Temporal Representation Vectors:} We provide a visualization of the temporal representation vectors concerning MCI and NC classes for each fold in Fig.~\ref{fig:embedding_visualization}, using T-SNE~\cite{vd2008visualizing}. More specifically, since the MLP module uses the temporal representation vectors associated with the sequence of sentences belonging to subjects within each class for the classification task, visualization of these vectors can provide an intuitive understanding of the discriminative power of the framework. As depicted in Fig.~\ref{fig:embedding_visualization}, the temporal representation vectors created by the SCA module for the MCI and NC classes are generally distinguishable, indicating that our proposed framework learned to generate distinct features from sequences of sentences for each class.

\subsubsection{Sequence-based Experimental Results}  \label{SEC_Evaluation_Experiments_Sequence_based}
As explained in Sec.~\ref{SEC_Methodology}, the proposed framework is trained to classify MCI and NC based on a sequence of sentences, with the final prediction being determined by averaging all the class-related probabilities. Thus, it is possible to assess the predictive performance of the proposed model for each sequence of sentences too. More clearly, since the input to our proposed framework is a sequence of sentences, we assess the performance of our framework at the sequence level to provide a more in-depth analysis. Given that not all sequences may contain perceptible features suitable for determining between MCI and NC, the assessment outcomes at the sequence level will likely be relatively lower, compared to the subject-based evaluations. This is because subject-based prediction takes into account the probability scores predicted by the proposed framework for all sequences of sentences related to a person, while for the sequence-based analysis, we evaluate the framework performance for each sequence of sentences solely.

As Table~\ref{tbl:tbl_acc_auc_seq} shows, the average accuracy and AUC for the sequence-level analysis are 72.45\%, and 72.39\%, respectively. Comparing the subject-based results presented in Table~\ref{tbl:tbl_acc_auc} with the corresponding sequence-based values in Table~\ref{tbl:tbl_acc_auc_seq} shows that neither accuracy nor AUC follow a consistent trend on a per-fold basis. To provide further clarity, regarding the subject-based accuracy, we observe the lowest values for both accuracy and AUC on fold 1, while for the sequence level, fold 1 achieves the two highest values of accuracy and AUC. These results can be interpreted as follows: For sequence-level analysis, we take into account the probabilities associated with both MCI and the NC classes and assign each sequence to the class with the higher probability. However, in subject-based analysis, we calculate the average of probabilities regarding each class for all sequences within a specific subject (see Sec.~\ref{SEC_Methodology}). 

Table~\ref{tbl:tbl_f1_pre_rec_seq} shows the sequence-level precision, recall, and F-1 score values associated with MCI and NC classes. For all provided metrics in Table~\ref{tbl:tbl_f1_pre_rec_seq}, the average performance of the framework is slightly higher for the NC class. Likewise, compared to similar values associated with the subject-level experiments in Table~\ref{tbl:tbl_f1_pre_rec_seq}, the sequence-level values show a relatively lower prediction performance, indicating that observing only one sequence might not provide adequate information for accurately distinguishing between MCI and NC classes.

\begin{figure*}[t]
  \centering
  \includegraphics[width=1.99\columnwidth]{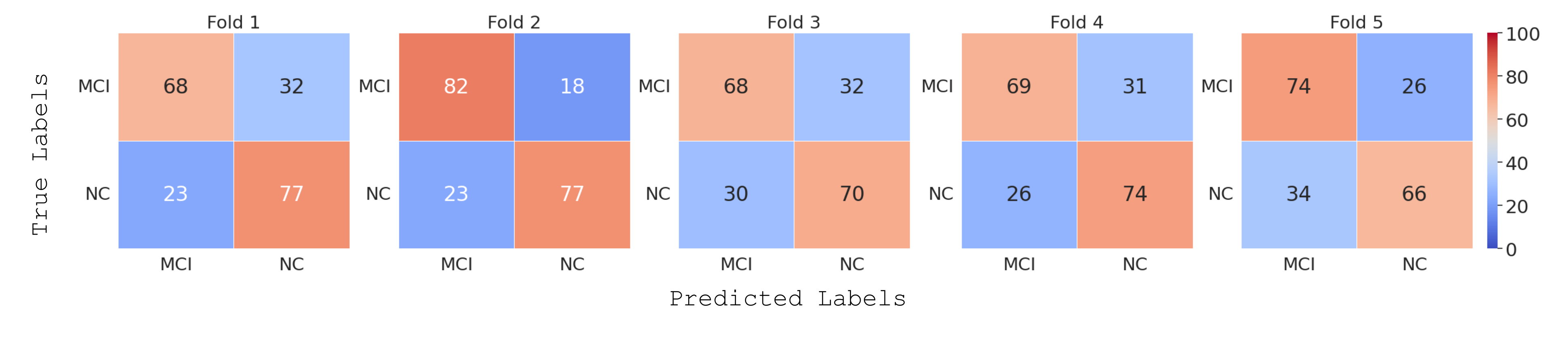}
  \caption{Sequence-level confusion matrices for each fold. All the values are presented in percentages.}
  \label{fig:conf_matrix_seq}
\end{figure*}

Fig.~\ref{fig:conf_matrix_seq} shows the sequence-based confusion matrices for each fold. For folds 1, 3, and 4, the sequence-level assessment shows that the framework performs better in correctly classifying the NC class, while for folds 2, and 5, the performance of the framework is better for the MCI class. 

\subsection{Comparison on I-CONECT}
In this section, we compare the result of our proposed framework with the state-of-the-art (SOTA) research on the I-CONECT dataset. Since I-CONECT does not have a designated training and test set, the previously proposed methods and research have designed their own training and test set. Consequently, although the comparison may not be entirely equitable due to the variations in dataset composition and configuration, it still offers a relative insight into the comparative performance of each research. Moreover, a majority of previously proposed methods and research have worked on some special \textit{themes} during both the training and testing phases, potentially limiting the generalizability of their approaches to specific thematic contexts. In contrast, our framework is theme-agnostic, aiming to extract features from sequences of sentences without dependence on specific themes, thus enhancing its potential for broader application and generalization.

Since accuracy is a less informative metric compared to the AUC, specifically on imbalanced datasets, we compare the performance of our framework, with the average of the AUC values mentioned in the SOTA research in Table~\ref{tbl:tbl_sota}. At a glance, it can be concluded from Table~\ref{tbl:tbl_sota} that the \textit{Visual} approaches achieved the lowest AUC values ( 56.02\%, and 79.25\% for Sun \etal~\cite{sun2023mc}, and Alsuhaibani \etal~\cite{alsuhaibani2023detection}, respectively), while the \textit{Linguistic} (and \textit{Acoustic \& Linguistic}) approaches achieved the highest AUC values. This outcome aligns with expectations, as language inherently encompasses a richer set of features compared to video alone when it comes to detecting MCI from the NC group. 

Regarding the Visual approaches, the method proposed by Alsuhaibani \etal~\cite{alsuhaibani2023detection} adopts a strategy where it generates embedded feature vectors for each frame and subsequently performs classification within the embedding space. This approach achieved a significantly higher AUC value (about 23.23\%) when compared to the method proposed by Sun \etal~\cite{sun2023mc} which directly uses images for the classification task. 

Considering \textit{Linguistic} approaches, our proposed framework achieved the AUC value of 85.16\%, which is the highest compared to the AUC values for the methods proposed by Asgari \etal~\cite{asgari2017predicting} (about 79.61\%), and Chen \etal~\cite{chen2020topic} (about 83.82\%). However, as the training and test set for the research mentioned in Table~\ref{tbl:tbl_sota} are not consistent, the comparison is \textbf{not} fair.

\begin{table}[t] 
\caption{The comparison between the average AUC value achieved by our framework and the values reported by the SOTA research on the I-CONECT dataset. }
\label{tbl:tbl_sota}
\centering
\small
\resizebox{0.45\textwidth}{!}
{{
\begin{tabular}{l|cc}
\hline
Method                             & Modality               & AUC (\%) \\ \hline
Asgari et al. (2017)~\cite{asgari2017predicting} - Relativity  & Linguistic             & 79.61  \\
Chen et al. (2020)~\cite{chen2020topic} - Table 2       & Linguistic             & 83.82  \\
Tang et al. (2022)~\cite{tang2022joint} - Table 2       & Acoustic \& Linguistic & 82.7   \\
Alsuhaibani et al. (2023)~\cite{alsuhaibani2023detection} - Table 4 & Visual                 & 79.25  \\
Sun et al. (2023)~\cite{sun2023mc} - Table 2        & Visual                 & 56.02  \\ \hdashline
ours                               & Linguistic             & 84.75  \\ \hline
\end{tabular}
}}
\end{table}

\subsection{Assessing Robustness Performance} \label{SEC_Evaluation_rb}
In the K-fold validation technique, the configuration of folds can impact the performance of the model. While this impact might be negligible, assessing the performance of the model using a group of different fold configurations, provides insights about the model generalization power. 

To this end, we repeat our 5-fold validation technique 5 times independently. For each fold configuration, we randomly create our 5 folds, train our model on each fold, and assess the performance on the corresponding test set. Table~\ref{tbl:tbl_gen_auc} shows the average AUC values for each fold configuration (the average AUC values for each fold within the corresponding fold configuration), as well as the overall average and std values for all 25 experiments. According to  Table~\ref{tbl:tbl_gen_auc}, the AUC values range between 74.99\% and 84.75\%. Likewise, the overall AUC of all 25 experiments is 79.94\%, with a std value of 1.93, showing that our proposed framework poses an acceptable generalization performance.

\begin{table}[t] 
\caption{The average AUC values of the proposed framework for each fold configuration. The average values and the standard deviation of all 5 fold configurations are provided.}
\label{tbl:tbl_gen_auc}
\centering
\small
\resizebox{0.45\textwidth}{!}
{{
\begin{tabular}{c|ccccc|cc}
\hline
Fold Configuration\# & 1 & 2 & 3 & 4 & 5 & Avg & std\\ \hline
AUC(\%)  & 84.75	&74.99	&78.33	&79.99	&81.66  & 79.94 & 1.93 \\ \hline
\end{tabular}
}}
\end{table}

\subsection{Ablation Study} \label{SEC_Evaluation_Ablation}
In this section, we investigate the effectiveness of various components of our proposed methodology such as InfoLoss, the hyper-parameter $\gamma$, and finally architecture of the SCA module. 

\subsubsection{Informative Loss vs Cross Entropy Loss} \label{SEC_Evaluation_Ablation_loss}
In this section, we present a comparison of the AUC values achieved by our proposed framework under two training conditions: one using the CE loss and the other utilizing InfoLoss. For further analysis, we provide both subject-level and sequence-level experimental results for each fold.

\begin{figure}[t!]
  \centering
  \includegraphics[width=0.90\columnwidth]{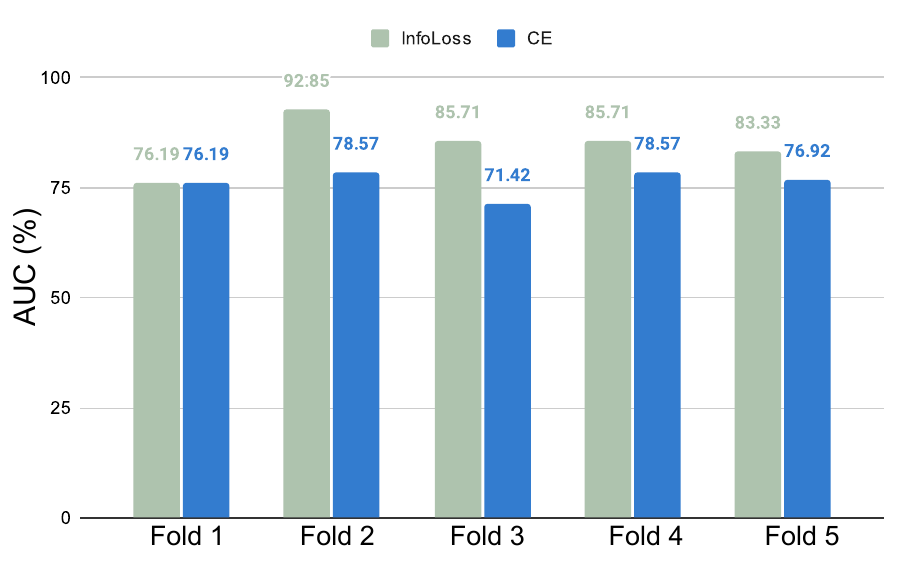}
  \caption{Subject-level comparison regarding AUC values achieved by utilizing InfoLoss compared to CE for each fold.}
  \label{fig:abl_loss_sub}
\end{figure}

Subject-level comparison of the AUC values regarding CE and InfoLoss is provided in Fig.~\ref{fig:abl_loss_sub}. As Fig.~\ref{fig:abl_loss_sub} shows, except for fold 1, the AUC values increase dramatically when the framework is trained using the InfoLoss. To be more detailed, for folds 2 and 3, we observe a significant increase in AUC values while training the framework with InfoLoss compared to CE, which are about 14.28\% (from 78.57\% to 92.85\%), and 14.29\% (from 71.42\% to 85.71\%) respectively. Likewise, for folds 4 and 5, the AUC values witness notable improvements, approximately 7.14\% (ascending from 78.57\% to 85.71\%), and 6.41\% (rising from 76.92\% to 83.33\%) respectively. Overall, In the subject-level experiments, the AUC exhibits an average increase of 8.42\% across all five folds when training the framework using InfoLoss.

\begin{figure}[t!]
  \centering
  \includegraphics[width=0.90\columnwidth]{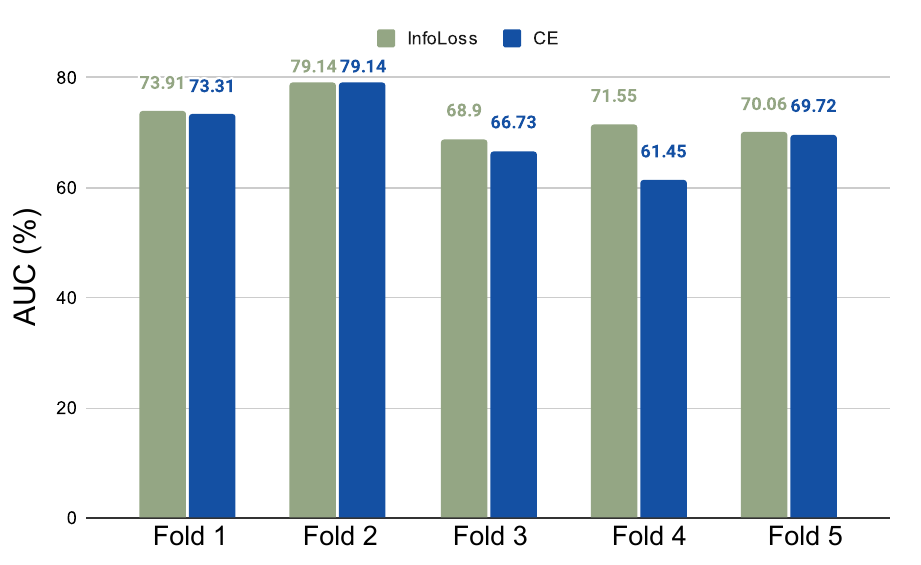}
  \caption{Sequence-level comparison regarding AUC values achieved by utilizing InfoLoss compared to CE for each fold.}
  \label{fig:abl_loss_sent}
\end{figure}

Fig.~\ref{fig:abl_loss_sent} presents a sequence-level comparison of the AUC values, between utilizing CE loss and InfoLoss. Overall, the AUC values achieved by utilizing InfoLoss are higher compared to those achieved with CE loss. Based on Fig.~\ref{fig:abl_loss_sent}, the least improvements occurred for fold 2, while the highest improvement is observed for fold 4. Furthermore, the average improvement of AUC for all 5 folds is about 2.62\%.

Additionally, a comparison between Fig.~\ref{fig:abl_loss_sub} and Fig.~\ref{fig:abl_loss_sent} reveals that the average AUC improvement achieved with InfoLoss over CE loss is much higher in the subject-level experiments. As discussed in Sec.~\ref{SEC_Methodology_CustomLoss_objective}, for subject-based comparison, we take the average of the probabilities regarding each class for all sequences belonging to a subject, and accordingly, label a subject with the class having greater probability. In contrast, for sequence-level analysis, we label a sequence with respect to the probabilities regarding each class. Hence, in subject-based analysis, the confidence (probabilities regarding each class) of each sequence plays a crucial role in the final classification of a subject. Having said that, it can be concluded by comparing the AUC values in Fig.~\ref{fig:abl_loss_sub} and Fig.~\ref{fig:abl_loss_sent} that using InfoLoss results in sequences with greater confidence compared to the CE. Hence, although the sequence-level comparison shows relatively similar performance between CE and InfoLoss, utilizing the latter results in a significant improvement in AUC value in subject-level comparison.

\subsubsection{Effect of the Hyper-Parameter $\gamma$} \label{SEC_Evaluation_Ablation_gamma}
The hyper-parameter $\gamma$ defines the number of sentences processed at each step by the framework. In Sec.~\ref{SEC_Methodology}, we define $\gamma=200$, indicating that each input to the framework is a sequence of 200 sentences (or phrases). In this section, we study how different values of this hyper-parameter can affect the accuracy of the classification.

Since our proposed framework is designed to learn the attention between a sequence of $\gamma$ sentences, we can conclude that setting $\gamma$ to a smaller value may intuitively lead to a decrease in the model prediction performance, whereas larger values of $\gamma$ are likely to result in higher classification rate, as the model can more efficiently extract more contextual nuances and relationships within longer sequences of sentences. To assess this hypothesis,  we conduct an empirical assessment by assigning 3 different values, which are 10, 100, and 200, to $\gamma$ and evaluate the framework's performance in terms of the AUC.

In the first experiment, we set $\gamma=10$, which means we fed the framework with 10 sequential sentences and performed the classification accordingly.  In this scenario, we observed that the training process was not stable and the model failed to converge. To conclude, the contextual nuances and the relationship between sentences within such short sequences may not be enough to effectively distinguish between the MCI and NC groups.

\begin{figure}[t]
  \centering
  \includegraphics[width=0.90\columnwidth]{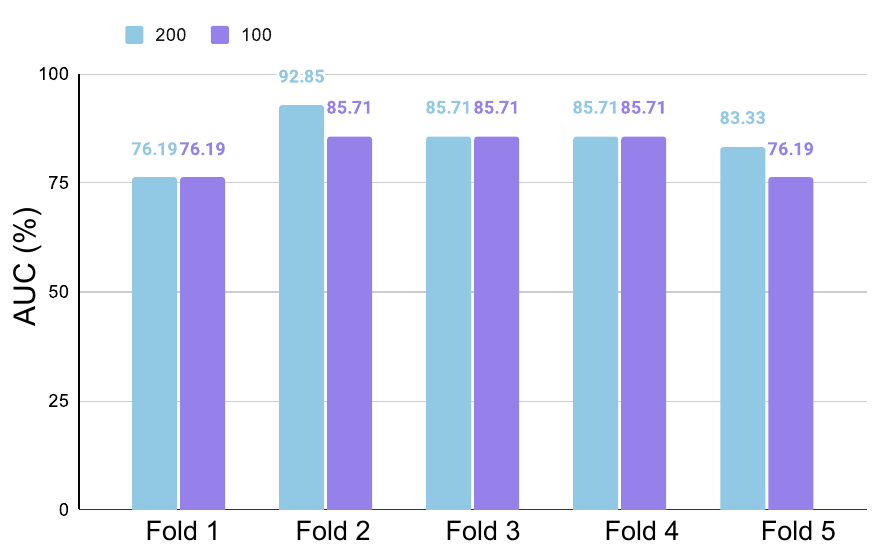}
  \caption{Subject-level comparison of AUC values with respect to different values of $\gamma$ for each fold.}
  \label{fig:abl_gamma_sub}
\end{figure}

\begin{figure}[t]
  \centering
  \includegraphics[width=0.90\columnwidth]{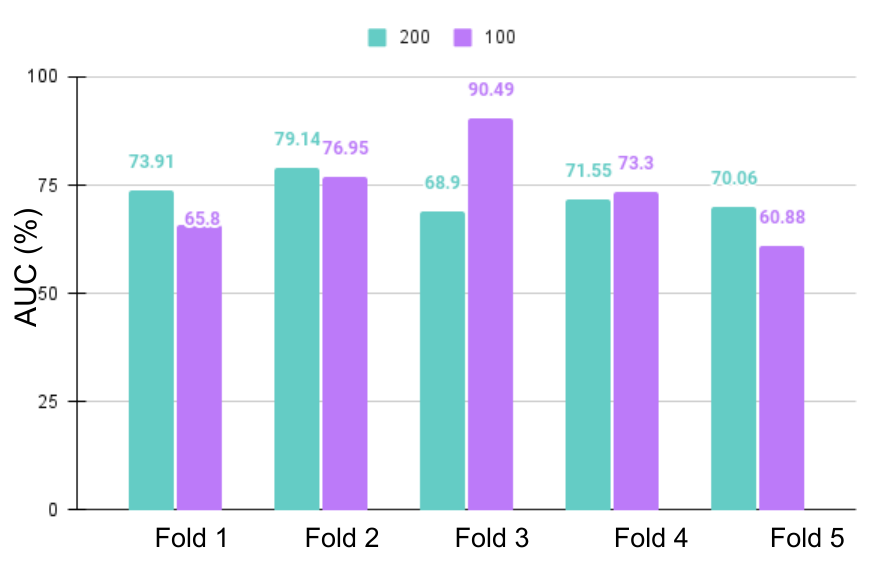}
  \caption{Sequence-level comparison of AUC values with respect to different values of $\gamma$ for each fold.}
  \label{fig:abl_gamma_sents}
\end{figure}

Fig.~\ref{fig:abl_gamma_sub} presents a comparison of per-fold AUC values for the subject-based evaluation of the framework. Overall, it can be concluded from Fig.~\ref{fig:abl_gamma_sub} that setting $\gamma=200$ results in better performance, while in most cases we observe equal performance. For a more comprehensive comparison, in Fig.~\ref{fig:abl_gamma_sents}, we provide a comparison of per-fold AUC values for the sequence-based evaluation of the framework. consequently, except for folds 3, and 4, the framework performs better in terms of the AUC, when $\gamma$ is set to 200 as compared to when it is set to 100. This result indicates that while the framework is more likely to extract additional features by processing longer sequences of sentences at once, there are examples where this extended sequence length may inadvertently lead to a reduction in the extraction of features that are particularly relevant for distinguishing between the MCI and NC classes. Hence, we set $\gamma=200$ for the evaluation of our framework in Sec.~\ref{SEC_Evaluation} as the best overall results are achieved accordingly.

\subsubsection{SCA Architecture} \label{SEC_Evaluation_Ablation_sca_arch}
As mentioned in Sec.~\ref{SEC_Methodology_SCA}, the SCA module is a Transformer encoder designed to learn the attention, and temporal features within a sequence of sentences to distinguish between the MCI and NC classes. Thus, the size of the fully connected layers affects the performance of our framework. Hence, we conducted an experiment to assess the effect of the size of the fully connected layers in the SCA module. 

\begin{figure}[t!]
  \centering
  \includegraphics[width=0.90\columnwidth]{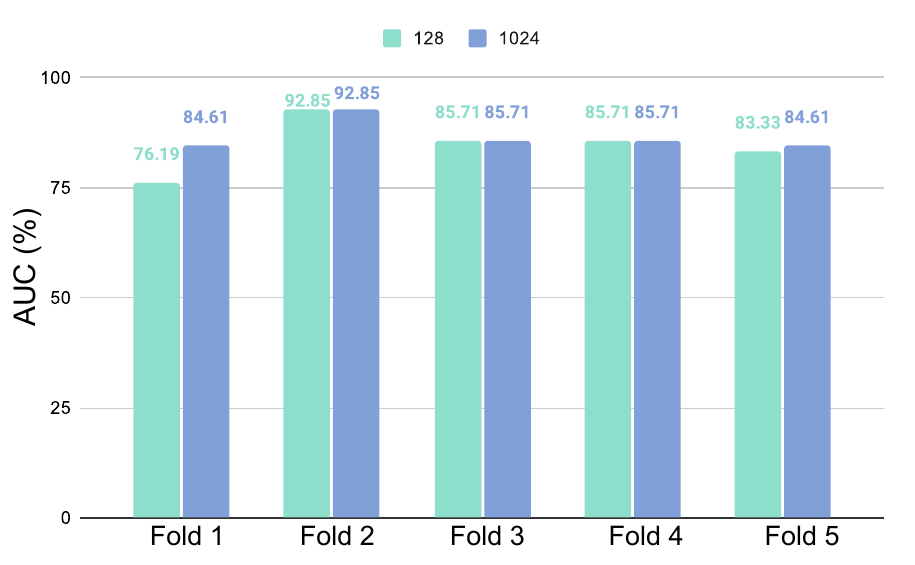}
  \caption{Subject-level comparison of AUC values across different sizes of fully connected layers in the SCA module for each fold.}
  \label{fig:abl_nn_size_sub}
\end{figure}

In Fig.~\ref{fig:abl_nn_size_sub}, we present a comparative analysis of the framework's performance (in terms of the AUC) regarding the size of the fully connected layers, for each fold, with a focus on subject-based evaluation. Accordingly, for the subject-based comparison, the proposed framework achieved a relatively similar AUC, except for fold 1, where the AUC increases from 76.19\% to 84.61\% when increasing the size of the fully connected layers from 128 to 1024. 

\begin{figure}[t!]
  \centering
  \includegraphics[width=0.90\columnwidth]{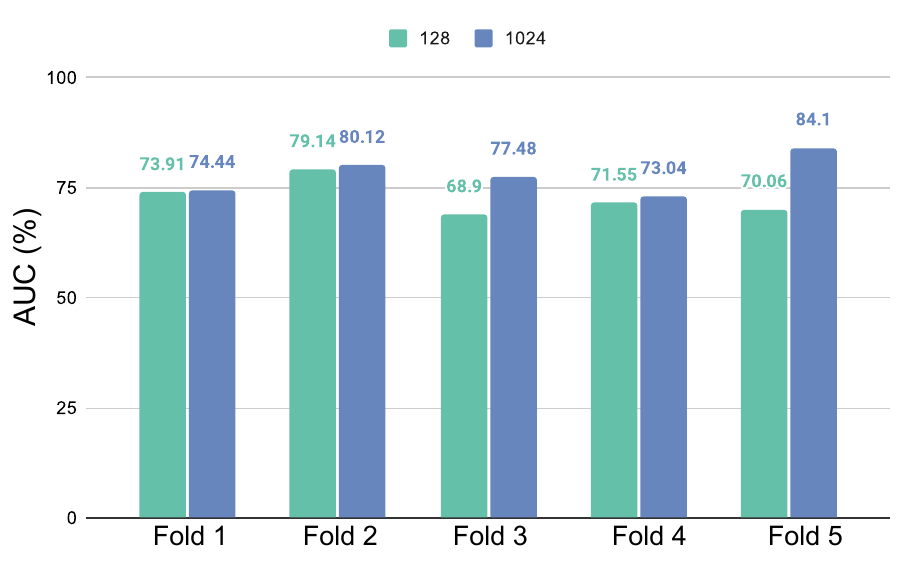}
  \caption{Sequence-level comparison of AUC values across different sizes of fully connected layers in the SCA module for each fold.}
  \label{fig:abl_nn_size_sent}
\end{figure}

For a more comprehensive comparative analysis, Fig.~\ref{fig:abl_nn_size_sent} presents the AUC results of the sequence-based experiments for each fold. These experiments consider two scenarios: one with the SCA module having fully connected layers of size 128 and the other with fully connected layers of size 1024. As Fig.~\ref{fig:abl_nn_size_sent} shows, overall, increasing the size of the fully connected layer from 128 to 1024 increases the sequence-level AUC of the framework. However, in most cases (folds 1,2, and 4) these improvements are about 1\%, which is negligible. 

While it can be concluded from Fig.~\ref{fig:abl_nn_size_sub}, and Fig.~\ref{fig:abl_nn_size_sent} that increasing the size of the fully connected layers in the SCA module has the potential to improve the discriminative power of the framework in distinguishing between MCI and NC classes, in most cases the improvement are relatively small. 

The number of Transformer encoders can indeed affect the performance of the proposed framework as well. However, utilizing more than one layer significantly increases the number of parameters within the framework, which can have implications for GPU memory limitations. Consequently, we were forced to reduce the batch size to approximately 20. Unfortunately, this adjustment led to instability during the training process, making the experiment unreliable for inclusion in the ablation study.

\begin{figure}[t]
  \centering
  \includegraphics[width=0.90\columnwidth]{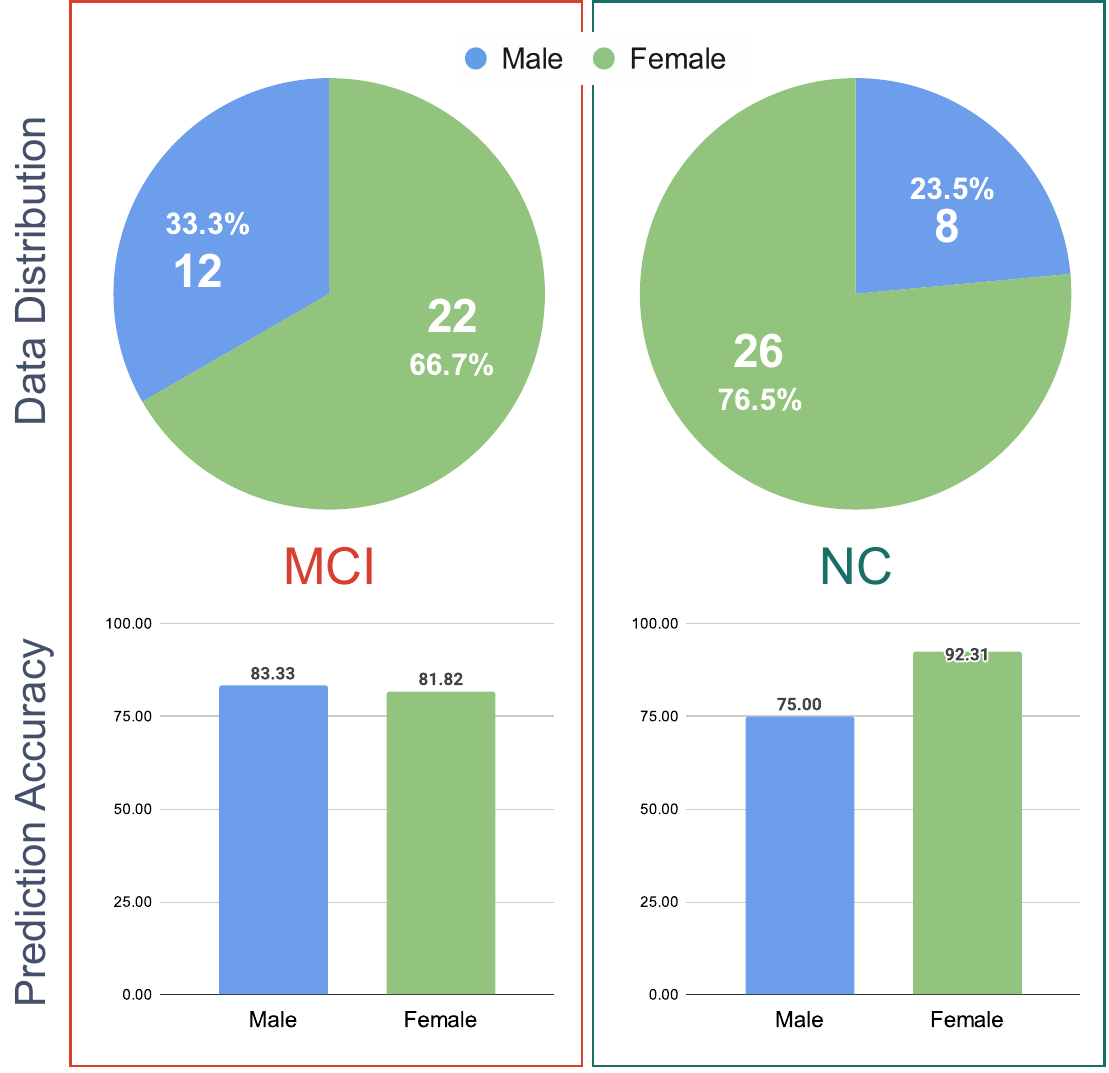}
  \caption{Gender distribution within the I-CONECT dataset among MCI and NC groups. The accuracy values show the portion of correctly classified subjects within each gender and the control group (MCI or NC).}
  \label{fig:gender_ana}
\end{figure}
%


\section{Discussion}  \label{SEC_Discussion}
Our in-depth analysis (see Sec.~\ref{SEC_Evaluation_Experiments}) of the performance of our proposed framework shows that in most cases, the framework performs more accurately in correctly detecting the NC group compared to the MCI. Since both the training and the test set are relatively balanced, it can be concluded that the better performance of the framework in detecting the NC group has not originated from the unbalanced data sources. Thus, we provide the following reasons that might explain this outcome. 

To begin with, as MCI is a complex and heterogeneous condition that is characterized by subtle cognitive deficits~\cite{langa2014diagnosis}, it might be hard to distinguish such impairment by analyzing language data solely. The intrinsic variability in MCI can make it challenging to capture explicit linguistic features from video transcripts alone. Furthermore, MCI-related cognitive decline can affect some cognitive domains more than others, resulting in variability in language impairments among individuals with MCI~\cite{petersen2004mild, sperling2011toward}. Thus, the subtle and context-dependent nature of MCI-related features can make them more challenging to identify, compared to the NC group.

\begin{figure}[t!]
  \centering
  \includegraphics[width=0.90\columnwidth]{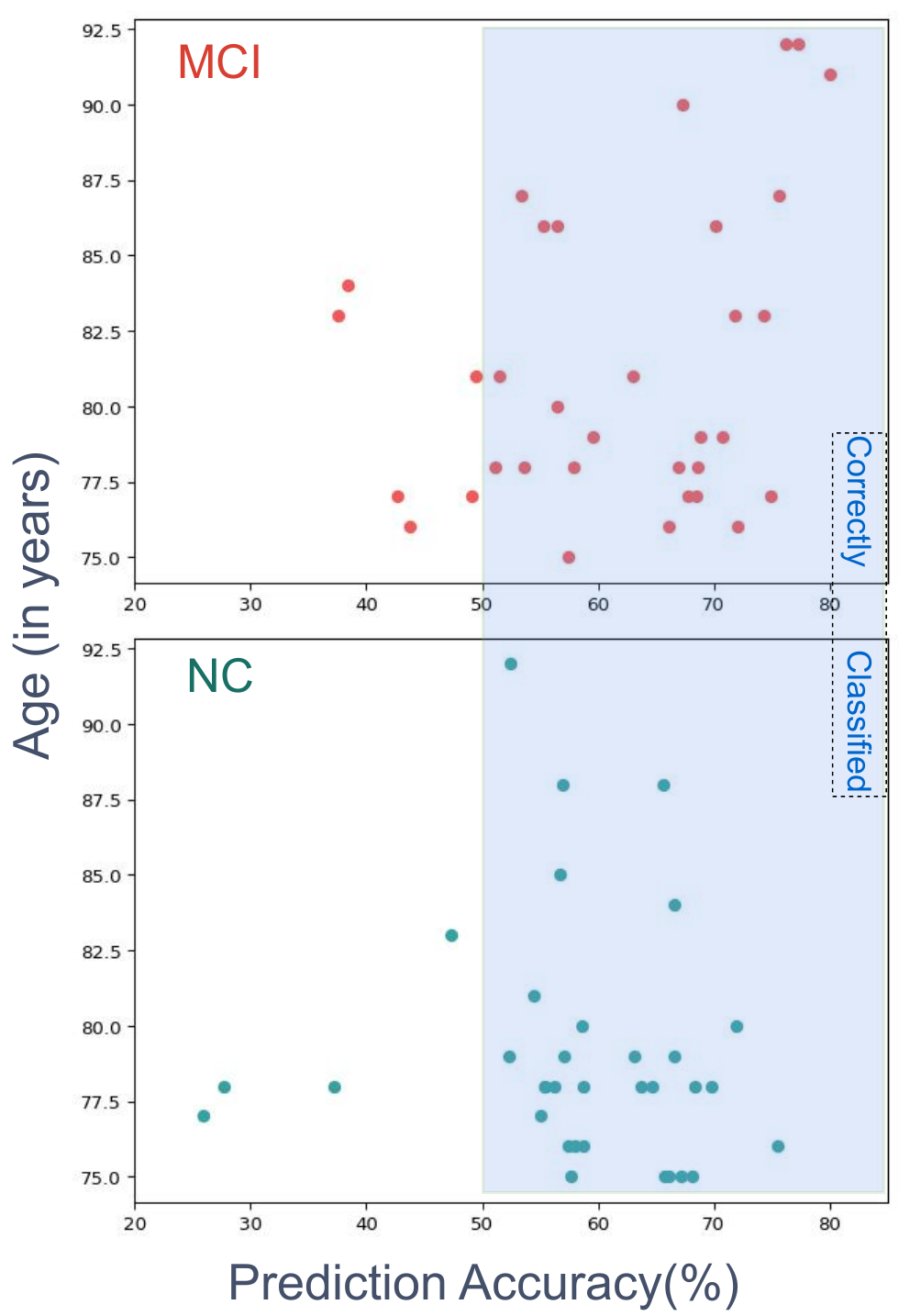}
  \caption{The age distribution (in years) in the I-CONECT dataset across MCI and NC groups. The prediction accuracy values represent the predicted probability scores for each subject within the I-CONECT dataset with respect to their corresponding class.}
  \label{fig:age_ana}
\end{figure}

The gender of subjects might impact the prediction accuracy within the MCI and NC group. Thus, In Fig.~\ref{fig:gender_ana}, we provided the gender distribution within the I-CONECT dataset among MCI and NC groups, alongside their prediction performance. As Fig.~\ref{fig:gender_ana} shows, for the MCI group, the prediction performance of our framework for both genders (female, and male) is relatively similar, while for the NC group, the framework shows much higher performance for the female group. It is important to note that the validity of this analysis is closely tied to the prediction performance of our proposed framework, and any extension of this analysis requires further investigations for broader generalization.

From another perspective, we can study the performance of the proposed framework with respect to the age of subjects. Hence, in Fig.~\ref{fig:age_ana}, we depicted the prediction accuracy of our proposed framework, with respect to the age of subjects. As Fig.~\ref{fig:age_ana} shows, the available subjects in the I-CONECT dataset age from about 75 to 92 years old. By examining Fig.~\ref{fig:age_ana}, we can make a preliminary inference that as the age of subjects increases, the number of correctly classified subjects also appears to increase. It is also remarkable that the number of subjects decreases as age increases, which can have an impact on the rate of correctly classified subjects.

\begin{figure}[t!]
  \centering
  \includegraphics[width=0.90\columnwidth]{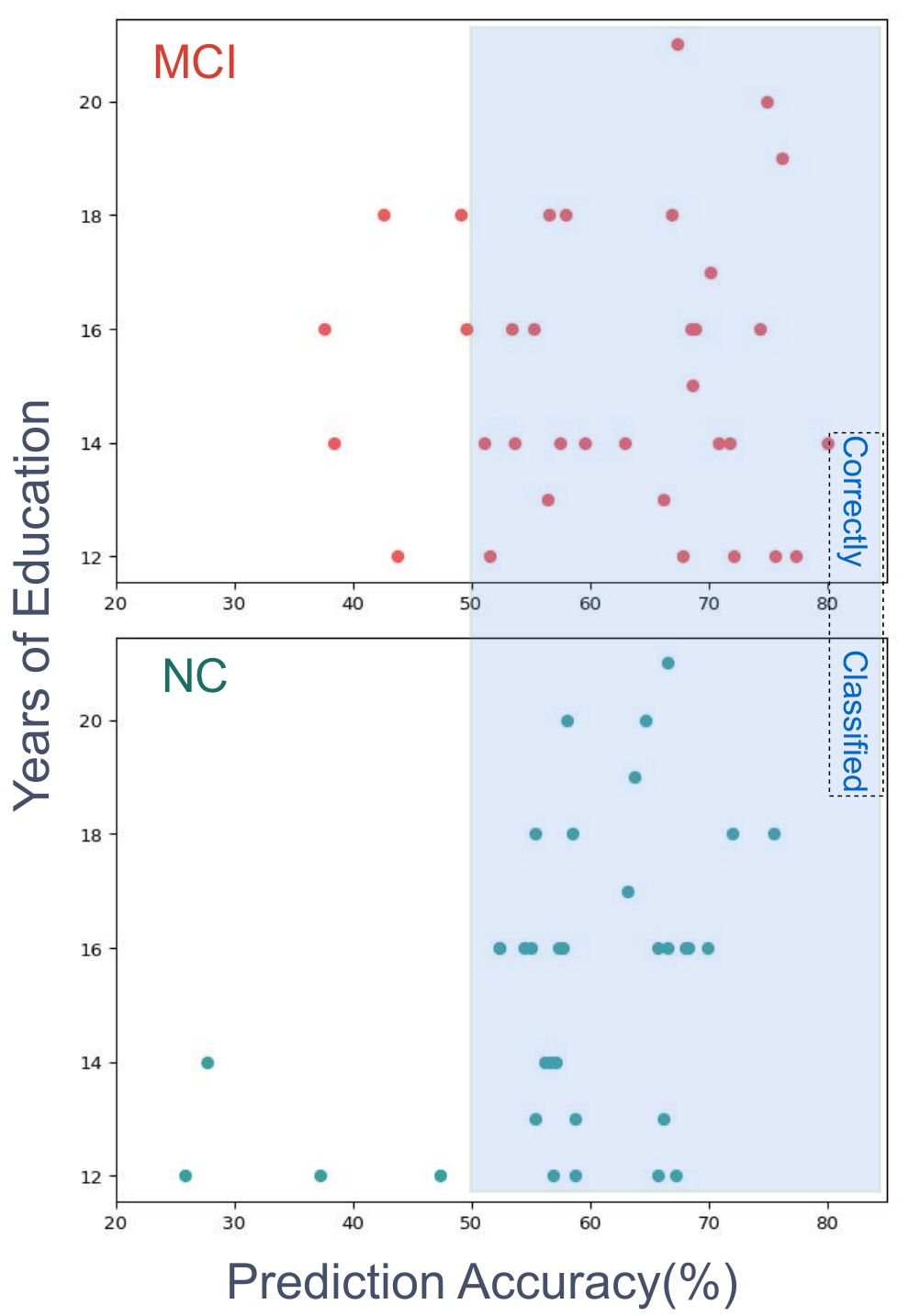}
  \caption{The education distribution (in years) in the I-CONECT dataset across MCI and NC groups. The prediction accuracy values represent the predicted probability scores for each subject within the I-CONECT dataset with respect to their corresponding class.}
  \label{fig:edu_ana}
\end{figure}

The education level of subjects might impact the prediction performance between MCI and NC groups. Thus, in Fig.~\ref{fig:edu_ana}, we depicted the prediction accuracy of our proposed framework, with respect to the years of education of each subject in the I-CONECT dataset. As Fig.~\ref{fig:edu_ana} shows, for the NC group, there is a correlation between the year of education and the prediction accuracy of our proposed framework, while for the MCI group, it is hard to find a meaningful pattern. Needless to say, expanding this conclusion requires conducting further investigations to confirm its generalization.

\section{Conclusion}  \label{SEC_Conclusion}
In this study, we proposed a Transformer-based framework for detecting between subjects with MCI and those with NC by analyzing the transcripts from video interviews conducted as part of the I-CONECT study project. More specifically, our proposed framework consists of two Transformer-based modules: SE and SCA modules. First, the SE module was applied to each sentence to capture the contextual relationships within words, which created the corresponding representation vector. Next, the SCA module was utilized to capture temporal features within the sequences of sentences, using the previously created sentential representation vectors. The output of the SCA module, which was the temporal representation vector associated with each sequence of sentences, was fed to an MLP for the classification of MCI and NC classes. To enhance the classification accuracy, we presented a novel loss function called InfoLoss, which took into account the uncertainty associated with the ground truth labels, particularly the number of sentences in each interview. 

To evaluate the effectiveness of our proposed framework, we conducted a comprehensive analysis considering a wide variety of evaluation metrics including accuracy, AUC, precision, recall, and F-1 score. Furthermore, we assessed the performance of our framework at the subject-level and sequence-level. In addition, we compared our results with the SOTA research on the I-CONECT dataset. In summary, our framework can detect MCI from NC with an AUC of 84.75\%.

In our future research plan, we will study how meta-data such as age, gender, and ethnicity will impact the model performance. Furthermore, we plan to extend the scope of our investigation by evaluating the accuracy and effectiveness of our proposed framework on individuals with AD.

\section*{Acknowledgement}
This research received partial support from the National Academy of Medicine Healthy Longevity Catalyst Award granted to DreamFace Technologies, LLC, the ECE Research Faculty Bridge Fund provided by the Dr. Frank Hielscher Gift, and funding from the National Institute of Health under grants R01AG051628 and R01AG056102.

\section*{Appendix}\label{SEC_Appendix}
As mentioned in Sec.~\ref{SEC_Methodology}, for each person $P^i$ in the I-CONECT dataset, there exists a set of video transcripts for that person, called $R^i$. The video transcripts are split into $\lambda$ number of sentences, called \textit{a sequence of sentences}, from $R^i$. For simplicity, we defined $Seq^i$, associated with $R^i$, as the set containing all sequences of sentences belonging to person $P^i$. Thus, for person $P^i$, we have $Seq:=\{seq_1, seq_2, ..., seq_k\}$. 

In Sec.~\ref{SEC_Methodology_CustomLoss_Freq}, we expressed that there is a reverse relationship between the number of sequences of sentences within transcripts associated with a specific subject and the amount of information achievable by observing each sequence. Put simply, we stated that the information gained (entropy reduction) from observing each sequence of sentences is inversely related to the total count of sequences associated with a subject. The subsequent section includes proof supporting our assertion.

\textbf{Proof:}
Assuming for person $P$ in the dataset, there exists $R$, a set of transcripts of video interviews. Based on our notation described in Sec.~\ref{SEC_Methodology_Formal}, $R$ contains $n$ sentences, and thereby, any $\lambda$ number of sequential sentences in $R$ is named a \textit{sequence of sentences}, and denoted as $seq_i$. More clearly, $seq_i$ contains sentences indexed from $i$ to $i+\lambda$. Hence, for person $P$, we have a set containing all sequences of sentences as $Seq:=\{ seq_1, seq_2, ..., seq_k\}$.

Now, assuming the class label for individual $P$ is unknown. We aim to infer the label by observing each element in the set of sequences, $Seq$. The class label associated with an arbitrary element of $Seq$ such as $seq_i$, can be either zero (for MCI) or one (for NC). Thus, $seq_i$ has a Bernoulli probability distribution, and accordingly, we have:
\begin{align}\label{eq:app_p_person}
\mathcal{P}(seq_i| MCI) =  \mathcal{P}(seq_i| NC) = 0.5
\end{align}
where $\mathcal{P}$ refers to \textit{Probability}. Also, these probability scores are independent. Intuitively, gradually observing each element of the $Seq$ set, gives us more confidence about the associated class label. By observing all sequences of sentences in the $Seq$ set, we can more confidently reveal the class label of person $P$.

Next, we calculate the entropy of the set of sequences, $Seq$ as follows in Eq.~\ref{eq:app_entropy_person}:
\begin{align}\label{eq:app_entropy_person}
H_0(Seq) =  - \sum_{i=1}^{k} \mathcal{P}(seq_i) log_2(\mathcal{P}(seq_i))
\end{align}
As $\mathcal{P}(seq_i) =0.5$, and $log_2(\mathcal{P}(seq_i)) = log_2(0.5)=-1$, we rewrite Eq.~\ref{eq:app_entropy_person}, as follow:
\begin{align}\label{eq:app_entropy_person_1}
H_0(Seq) =  - \sum_{i=1}^{k} 0.5 \times -1 = \frac{k}{2}
\end{align}

Now, let's assume we have observed the first sequence of sentences $seq_1$ in $Seq$. We calculate the entropy of the set of sequences, $Seq$, after this \textit{observation}, as follows in Eq.~\ref{eq:app_entropy_person_obs}:
\begin{align}\label{eq:app_entropy_person_obs}
\begin{split} 
H_1(Seq) =  & - (\sum_{i=2}^{k} \mathcal{P}(seq_i) log_2(\mathcal{P}(seq_i)) + \\
 & \mathcal{P}(seq_1) log_2(\mathcal{P}(seq_1)))
\end{split}
\end{align}
As we have observed the label of $seq_1$, we can now write $\mathcal{P}(seq_1) = 1$, and $log_2(1) = 0$. Thus, we summarize Eq.~\ref{eq:app_entropy_person_obs} as follows:
\begin{align}\label{eq:app_entropy_person_obs_1}
\begin{split} 
H_1(Seq) =  & \frac{k-1}{2} + 0 = \frac{k-1}{2} 
\end{split}
\end{align}
Comparing the outcome of Eq.~\ref{eq:app_entropy_person_1}, and Eq.~\ref{eq:app_entropy_person_obs_1}, we reach a conclusion that observing a sequence of sentences $seq_1$, from $Seq$ result in a reduction in the entropy of $Seq$, and hence, provide more \textit{information} regarding the class label of person $P$. 

Now, we can calculate the \textit{portion of the reduction} in entropy regarding $Seq$ after observing one sequence, as follows in Eq.~\ref{eq:app_entropy_person_obs}:
\begin{align}\label{eq:app_entropy_person_obs}
\begin{split}
H_{reduction} = & \frac{\frac{k}{2} - (\frac{k-1}{2})}{\frac{k}{2}} = 
1- (\frac{\frac{k-1}{2})}{\frac{k}{2}}) = \\
& 1-(1-\frac{1}{k}) = \frac{1}{k}
\end{split}
\end{align}
The result of Eq.~\ref{eq:app_entropy_person_obs} shows that the reduction in the entropy for a set $Seq$ containing $k$ sequences of sentences, after observing one arbitrary sequence, $seq_i$, is inversely proportional to the $k$.

\textbf{Example:} For clarification, consider a dataset with two individuals, $P^1$ and $P^2$. The number of sentences available for $P^1$ is 100, and for $P^2$ it is 1000. Let $\lambda = 10$. Hence, the length of each person's sequence of sentences is 10. Consequently, we have $Seq^1:=\{ seq^1_0, ..., seq^1_9\}$ and $Seq^2:=\{ seq^2_0, ..., seq^2_{99}\}$. The length of $Seq^1$ is denoted as $k^1=10$, and for $Seq^2$, it is represented as $k^2=100$. Utilizing Eq.~\ref{eq:app_entropy_person_1}, the entropy of $Seq^1$ is $H(Seq^1) = \frac{10}{2}=5$, and entropy for $Seq^2$ is $H(Seq^2) = \frac{100}{2}=50$. Now, we observe one sample (one sequence of sentences) from each set, $Seq^1$ and $Seq^2$. Using Eq.~\ref{eq:app_entropy_person_obs}, the reduction in the entropy of $Seq^1$ is $\frac{1}{10}$, while the reduction in the entropy of $Seq^2$ is $\frac{1}{100}$.

This example demonstrates an intuitive concept as follows: for a person with fewer transcripts, each sentence (and consequently each sequence of sentences) provides more information about the class label, in contrast to a sentence (or a sequence of sentences) from a subject with a higher number of transcripts.

\printcredits

\section*{Conflict of interest statement}
The authors declare that they have no known competing financial interests or personal relationships that could have appeared to influence the work reported in this paper.

\bibliographystyle{cas-model2-names}

\bibliography{ref}

\begin{thebibliography}{68}
\expandafter\ifx\csname natexlab\endcsname\relax\def\natexlab#1{#1}\fi
\providecommand{\url}[1]{\texttt{#1}}
\providecommand{\href}[2]{#2}
\providecommand{\path}[1]{#1}
\providecommand{\DOIprefix}{doi:}
\providecommand{\ArXivprefix}{arXiv:}
\providecommand{\URLprefix}{URL: }
\providecommand{\Pubmedprefix}{pmid:}
\providecommand{\doi}[1]{\href{http://dx.doi.org/#1}{\path{#1}}}
\providecommand{\Pubmed}[1]{\href{pmid:#1}{\path{#1}}}
\providecommand{\bibinfo}[2]{#2}
\ifx\xfnm\relax \def\xfnm[#1]{\unskip,\space#1}\fi
\bibitem[{Alsuhaibani et~al.(2023)Alsuhaibani, Dodge and Mahoor}]{alsuhaibani2023detection}
\bibinfo{author}{Alsuhaibani, M.}, \bibinfo{author}{Dodge, H.H.}, \bibinfo{author}{Mahoor, M.H.}, \bibinfo{year}{2023}.
\newblock \bibinfo{title}{Detection of mild cognitive impairment using facial features in video conversations}.
\newblock \bibinfo{journal}{arXiv preprint arXiv:2308.15624} .
\bibitem[{{Alzheimer's~and~Dementia}(2023)}]{alzWhatAlzheimers}
\bibinfo{author}{{Alzheimer's~and~Dementia}}, \bibinfo{year}{2023}.
\newblock \bibinfo{title}{What is alzheimer's?}
\newblock \bibinfo{howpublished}{\url{https://www.alz.org/alzheimers-dementia/what-is-alzheimers}}.
\newblock \bibinfo{note}{[Accessed 21-08-2023]}.
\bibitem[{Amini et~al.(2023)Amini, Hao, Zhang, Song, Gupta, Karjadi, Kolachalama, Au and Paschalidis}]{amini2023automated}
\bibinfo{author}{Amini, S.}, \bibinfo{author}{Hao, B.}, \bibinfo{author}{Zhang, L.}, \bibinfo{author}{Song, M.}, \bibinfo{author}{Gupta, A.}, \bibinfo{author}{Karjadi, C.}, \bibinfo{author}{Kolachalama, V.B.}, \bibinfo{author}{Au, R.}, \bibinfo{author}{Paschalidis, I.C.}, \bibinfo{year}{2023}.
\newblock \bibinfo{title}{Automated detection of mild cognitive impairment and dementia from voice recordings: A natural language processing approach}.
\newblock \bibinfo{journal}{Alzheimer's \& Dementia} \bibinfo{volume}{19}, \bibinfo{pages}{946--955}.
\bibitem[{Arnab et~al.(2021)Arnab, Dehghani, Heigold, Sun, Lu{\v{c}}i{\'c} and Schmid}]{arnab2021vivit}
\bibinfo{author}{Arnab, A.}, \bibinfo{author}{Dehghani, M.}, \bibinfo{author}{Heigold, G.}, \bibinfo{author}{Sun, C.}, \bibinfo{author}{Lu{\v{c}}i{\'c}, M.}, \bibinfo{author}{Schmid, C.}, \bibinfo{year}{2021}.
\newblock \bibinfo{title}{Vivit: A video vision transformer}, in: \bibinfo{booktitle}{Proceedings of the IEEE/CVF international conference on computer vision}, pp. \bibinfo{pages}{6836--6846}.
\bibitem[{Asgari et~al.(2017)Asgari, Kaye and Dodge}]{asgari2017predicting}
\bibinfo{author}{Asgari, M.}, \bibinfo{author}{Kaye, J.}, \bibinfo{author}{Dodge, H.}, \bibinfo{year}{2017}.
\newblock \bibinfo{title}{Predicting mild cognitive impairment from spontaneous spoken utterances}.
\newblock \bibinfo{journal}{Alzheimer's \& Dementia: Translational Research \& Clinical Interventions} \bibinfo{volume}{3}, \bibinfo{pages}{219--228}.
\bibitem[{Becker et~al.(1994)Becker, Boiler, Lopez, Saxton and McGonigle}]{becker1994natural}
\bibinfo{author}{Becker, J.T.}, \bibinfo{author}{Boiler, F.}, \bibinfo{author}{Lopez, O.L.}, \bibinfo{author}{Saxton, J.}, \bibinfo{author}{McGonigle, K.L.}, \bibinfo{year}{1994}.
\newblock \bibinfo{title}{The natural history of alzheimer's disease: description of study cohort and accuracy of diagnosis}.
\newblock \bibinfo{journal}{Archives of neurology} \bibinfo{volume}{51}, \bibinfo{pages}{585--594}.
\bibitem[{Bertini et~al.(2022)Bertini, Allevi, Lutero, Calz{\`a} and Montesi}]{bertini2022automatic}
\bibinfo{author}{Bertini, F.}, \bibinfo{author}{Allevi, D.}, \bibinfo{author}{Lutero, G.}, \bibinfo{author}{Calz{\`a}, L.}, \bibinfo{author}{Montesi, D.}, \bibinfo{year}{2022}.
\newblock \bibinfo{title}{An automatic alzheimer’s disease classifier based on spontaneous spoken english}.
\newblock \bibinfo{journal}{Computer Speech \& Language} \bibinfo{volume}{72}, \bibinfo{pages}{101298}.
\bibitem[{Boschi et~al.(2017)Boschi, Catrical{\`a}, Consonni, Chesi, Moro and Cappa}]{boschi2017connected}
\bibinfo{author}{Boschi, V.}, \bibinfo{author}{Catrical{\`a}, E.}, \bibinfo{author}{Consonni, M.}, \bibinfo{author}{Chesi, C.}, \bibinfo{author}{Moro, A.}, \bibinfo{author}{Cappa, S.}, \bibinfo{year}{2017}.
\newblock \bibinfo{title}{Connected speech in neurodegenerative language disorders: a review. front psychol 6 (8): 269}.
\bibitem[{Brown et~al.(2020)Brown, Mann, Ryder, Subbiah, Kaplan, Dhariwal, Neelakantan, Shyam, Sastry, Askell et~al.}]{brown2020language}
\bibinfo{author}{Brown, T.}, \bibinfo{author}{Mann, B.}, \bibinfo{author}{Ryder, N.}, \bibinfo{author}{Subbiah, M.}, \bibinfo{author}{Kaplan, J.D.}, \bibinfo{author}{Dhariwal, P.}, \bibinfo{author}{Neelakantan, A.}, \bibinfo{author}{Shyam, P.}, \bibinfo{author}{Sastry, G.}, \bibinfo{author}{Askell, A.}, et~al., \bibinfo{year}{2020}.
\newblock \bibinfo{title}{Language models are few-shot learners}.
\newblock \bibinfo{journal}{Advances in neural information processing systems} \bibinfo{volume}{33}, \bibinfo{pages}{1877--1901}.
\bibitem[{Calz{\`a} et~al.(2021)Calz{\`a}, Gagliardi, Favretti and Tamburini}]{calza2021linguistic}
\bibinfo{author}{Calz{\`a}, L.}, \bibinfo{author}{Gagliardi, G.}, \bibinfo{author}{Favretti, R.R.}, \bibinfo{author}{Tamburini, F.}, \bibinfo{year}{2021}.
\newblock \bibinfo{title}{Linguistic features and automatic classifiers for identifying mild cognitive impairment and dementia}.
\newblock \bibinfo{journal}{Computer Speech \& Language} \bibinfo{volume}{65}, \bibinfo{pages}{101113}.
\bibitem[{Camgoz et~al.(2020)Camgoz, Koller, Hadfield and Bowden}]{camgoz2020sign}
\bibinfo{author}{Camgoz, N.C.}, \bibinfo{author}{Koller, O.}, \bibinfo{author}{Hadfield, S.}, \bibinfo{author}{Bowden, R.}, \bibinfo{year}{2020}.
\newblock \bibinfo{title}{Sign language transformers: Joint end-to-end sign language recognition and translation}, in: \bibinfo{booktitle}{Proceedings of the IEEE/CVF conference on computer vision and pattern recognition}, pp. \bibinfo{pages}{10023--10033}.
\bibitem[{Chen and Asgari(2021)}]{chen2021refining}
\bibinfo{author}{Chen, L.}, \bibinfo{author}{Asgari, M.}, \bibinfo{year}{2021}.
\newblock \bibinfo{title}{Refining automatic speech recognition system for older adults}, in: \bibinfo{booktitle}{ICASSP 2021-2021 IEEE International Conference on Acoustics, Speech and Signal Processing (ICASSP)}, \bibinfo{organization}{IEEE}. pp. \bibinfo{pages}{7003--7007}.
\bibitem[{Chen et~al.(2020)Chen, Dodge and Asgari}]{chen2020topic}
\bibinfo{author}{Chen, L.}, \bibinfo{author}{Dodge, H.H.}, \bibinfo{author}{Asgari, M.}, \bibinfo{year}{2020}.
\newblock \bibinfo{title}{Topic-based measures of conversation for detecting mild cognitive impairment}, in: \bibinfo{booktitle}{Proceedings of the conference. Association for Computational Linguistics. Meeting}, \bibinfo{organization}{NIH Public Access}. p.~\bibinfo{pages}{63}.
\bibitem[{Clarke et~al.(2021)Clarke, Barrick and Garrard}]{clarke2021comparison}
\bibinfo{author}{Clarke, N.}, \bibinfo{author}{Barrick, T.R.}, \bibinfo{author}{Garrard, P.}, \bibinfo{year}{2021}.
\newblock \bibinfo{title}{A comparison of connected speech tasks for detecting early alzheimer’s disease and mild cognitive impairment using natural language processing and machine learning}.
\newblock \bibinfo{journal}{Frontiers in Computer Science} \bibinfo{volume}{3}, \bibinfo{pages}{634360}.
\bibitem[{Colla et~al.(2022)Colla, Delsanto, Agosto, Vitiello and Radicioni}]{colla2022semantic}
\bibinfo{author}{Colla, D.}, \bibinfo{author}{Delsanto, M.}, \bibinfo{author}{Agosto, M.}, \bibinfo{author}{Vitiello, B.}, \bibinfo{author}{Radicioni, D.P.}, \bibinfo{year}{2022}.
\newblock \bibinfo{title}{Semantic coherence markers: The contribution of perplexity metrics}.
\newblock \bibinfo{journal}{Artificial Intelligence in Medicine} \bibinfo{volume}{134}, \bibinfo{pages}{102393}.
\bibitem[{Degottex et~al.(2014)Degottex, Kane, Drugman, Raitio and Scherer}]{degottex2014covarep}
\bibinfo{author}{Degottex, G.}, \bibinfo{author}{Kane, J.}, \bibinfo{author}{Drugman, T.}, \bibinfo{author}{Raitio, T.}, \bibinfo{author}{Scherer, S.}, \bibinfo{year}{2014}.
\newblock \bibinfo{title}{Covarep—a collaborative voice analysis repository for speech technologies}, in: \bibinfo{booktitle}{2014 ieee international conference on acoustics, speech and signal processing (icassp)}, \bibinfo{organization}{IEEE}. pp. \bibinfo{pages}{960--964}.
\bibitem[{Devlin et~al.(2018)Devlin, Chang, Lee and Toutanova}]{devlin2018bert}
\bibinfo{author}{Devlin, J.}, \bibinfo{author}{Chang, M.W.}, \bibinfo{author}{Lee, K.}, \bibinfo{author}{Toutanova, K.}, \bibinfo{year}{2018}.
\newblock \bibinfo{title}{Bert: Pre-training of deep bidirectional transformers for language understanding}.
\newblock \bibinfo{journal}{arXiv preprint arXiv:1810.04805} .
\bibitem[{Dosovitskiy et~al.(2020)Dosovitskiy, Beyer, Kolesnikov, Weissenborn, Zhai, Unterthiner, Dehghani, Minderer, Heigold, Gelly et~al.}]{dosovitskiy2020image}
\bibinfo{author}{Dosovitskiy, A.}, \bibinfo{author}{Beyer, L.}, \bibinfo{author}{Kolesnikov, A.}, \bibinfo{author}{Weissenborn, D.}, \bibinfo{author}{Zhai, X.}, \bibinfo{author}{Unterthiner, T.}, \bibinfo{author}{Dehghani, M.}, \bibinfo{author}{Minderer, M.}, \bibinfo{author}{Heigold, G.}, \bibinfo{author}{Gelly, S.}, et~al., \bibinfo{year}{2020}.
\newblock \bibinfo{title}{An image is worth 16x16 words: Transformers for image recognition at scale}.
\newblock \bibinfo{journal}{arXiv preprint arXiv:2010.11929} .
\bibitem[{Fader et~al.(2014)Fader, Zettlemoyer and Etzioni}]{fader2014open}
\bibinfo{author}{Fader, A.}, \bibinfo{author}{Zettlemoyer, L.}, \bibinfo{author}{Etzioni, O.}, \bibinfo{year}{2014}.
\newblock \bibinfo{title}{Open question answering over curated and extracted knowledge bases}, in: \bibinfo{booktitle}{Proceedings of the 20th ACM SIGKDD international conference on Knowledge discovery and data mining}, pp. \bibinfo{pages}{1156--1165}.
\bibitem[{Fard et~al.(2021)Fard, Abdollahi and Mahoor}]{fard2021asmnet}
\bibinfo{author}{Fard, A.P.}, \bibinfo{author}{Abdollahi, H.}, \bibinfo{author}{Mahoor, M.}, \bibinfo{year}{2021}.
\newblock \bibinfo{title}{Asmnet: A lightweight deep neural network for face alignment and pose estimation}, in: \bibinfo{booktitle}{Proceedings of the IEEE/CVF Conference on computer vision and pattern recognition}, pp. \bibinfo{pages}{1521--1530}.
\bibitem[{Fard et~al.(2022)Fard, Ferrantelli, Dupuis and Mahoor}]{fard2022sagittal}
\bibinfo{author}{Fard, A.P.}, \bibinfo{author}{Ferrantelli, J.}, \bibinfo{author}{Dupuis, A.L.}, \bibinfo{author}{Mahoor, M.H.}, \bibinfo{year}{2022}.
\newblock \bibinfo{title}{Sagittal cervical spine landmark point detection in x-ray using deep convolutional neural networks}.
\newblock \bibinfo{journal}{IEEE Access} \bibinfo{volume}{10}, \bibinfo{pages}{59413--59427}.
\bibitem[{Fard and Mahoor(2022a)}]{fard2022acr}
\bibinfo{author}{Fard, A.P.}, \bibinfo{author}{Mahoor, M.H.}, \bibinfo{year}{2022}a.
\newblock \bibinfo{title}{Acr loss: Adaptive coordinate-based regression loss for face alignment}, in: \bibinfo{booktitle}{2022 26th International Conference on Pattern Recognition (ICPR)}, \bibinfo{organization}{IEEE}. pp. \bibinfo{pages}{1807--1814}.
\bibitem[{Fard and Mahoor(2022b)}]{fard2022ad}
\bibinfo{author}{Fard, A.P.}, \bibinfo{author}{Mahoor, M.H.}, \bibinfo{year}{2022}b.
\newblock \bibinfo{title}{Ad-corre: Adaptive correlation-based loss for facial expression recognition in the wild}.
\newblock \bibinfo{journal}{IEEE Access} \bibinfo{volume}{10}, \bibinfo{pages}{26756--26768}.
\bibitem[{Fard and Mahoor(2022c)}]{fard2022facial}
\bibinfo{author}{Fard, A.P.}, \bibinfo{author}{Mahoor, M.H.}, \bibinfo{year}{2022}c.
\newblock \bibinfo{title}{Facial landmark points detection using knowledge distillation-based neural networks}.
\newblock \bibinfo{journal}{Computer Vision and Image Understanding} \bibinfo{volume}{215}, \bibinfo{pages}{103316}.
\bibitem[{Fard et~al.(2023)Fard, Mahoor, Lamer and Sweeny}]{fard2023ganalyzer}
\bibinfo{author}{Fard, A.P.}, \bibinfo{author}{Mahoor, M.H.}, \bibinfo{author}{Lamer, S.A.}, \bibinfo{author}{Sweeny, T.}, \bibinfo{year}{2023}.
\newblock \bibinfo{title}{Ganalyzer: Analysis and manipulation of gans latent space for controllable face synthesis}.
\newblock \bibinfo{journal}{arXiv preprint arXiv:2302.00908} .
\bibitem[{Ferris and Farlow(2013)}]{ferris2013language}
\bibinfo{author}{Ferris, S.H.}, \bibinfo{author}{Farlow, M.}, \bibinfo{year}{2013}.
\newblock \bibinfo{title}{Language impairment in alzheimer’s disease and benefits of acetylcholinesterase inhibitors}.
\newblock \bibinfo{journal}{Clinical interventions in aging} , \bibinfo{pages}{1007--1014}.
\bibitem[{Fraser et~al.(2016)Fraser, Meltzer and Rudzicz}]{fraser2016linguistic}
\bibinfo{author}{Fraser, K.C.}, \bibinfo{author}{Meltzer, J.A.}, \bibinfo{author}{Rudzicz, F.}, \bibinfo{year}{2016}.
\newblock \bibinfo{title}{Linguistic features identify alzheimer’s disease in narrative speech}.
\newblock \bibinfo{journal}{Journal of Alzheimer's Disease} \bibinfo{volume}{49}, \bibinfo{pages}{407--422}.
\bibitem[{Gilles(2022)}]{gilles2022age}
\bibinfo{author}{Gilles, C.}, \bibinfo{year}{2022}.
\newblock \bibinfo{title}{Age-related mild cognitive deficit: a ready-to-use concept?}
\newblock \bibinfo{journal}{Dialogues in Clinical Neuroscience} .
\bibitem[{Gottschalk and Bechtel(1989)}]{gottschalk1989computerized}
\bibinfo{author}{Gottschalk, L.A.}, \bibinfo{author}{Bechtel, R.J.}, \bibinfo{year}{1989}.
\newblock \bibinfo{title}{Computerized content analysis of natural language}.
\newblock \bibinfo{journal}{Artificial Intelligence in Medicine} \bibinfo{volume}{1}, \bibinfo{pages}{131--137}.
\bibitem[{Henderson et~al.(2019)Henderson, Budzianowski, Casanueva, Coope, Gerz, Kumar, Mrk{\v{s}}i{\'c}, Spithourakis, Su, Vuli{\'c} et~al.}]{henderson2019repository}
\bibinfo{author}{Henderson, M.}, \bibinfo{author}{Budzianowski, P.}, \bibinfo{author}{Casanueva, I.}, \bibinfo{author}{Coope, S.}, \bibinfo{author}{Gerz, D.}, \bibinfo{author}{Kumar, G.}, \bibinfo{author}{Mrk{\v{s}}i{\'c}, N.}, \bibinfo{author}{Spithourakis, G.}, \bibinfo{author}{Su, P.H.}, \bibinfo{author}{Vuli{\'c}, I.}, et~al., \bibinfo{year}{2019}.
\newblock \bibinfo{title}{A repository of conversational datasets}.
\newblock \bibinfo{journal}{arXiv preprint arXiv:1904.06472} .
\bibitem[{HuggingFace()}]{huggingfaceSentencetransformersallmpnetbasev2Hugging}
\bibinfo{author}{HuggingFace}, .
\newblock \bibinfo{title}{Sentence transformers}.
\newblock \bibinfo{howpublished}{\url{https://huggingface.co/sentence-transformers/all-mpnet-base-v2}}.
\newblock \bibinfo{note}{[Accessed 24-08-2023]}.
\bibitem[{Hussein et~al.(2022)Hussein, Chan, Van~Vleck, Beers, Mindt, Wolf, Curtis, Agarwal, Wisnivesky, Nadkarni et~al.}]{hussein2022natural}
\bibinfo{author}{Hussein, K.I.}, \bibinfo{author}{Chan, L.}, \bibinfo{author}{Van~Vleck, T.}, \bibinfo{author}{Beers, K.}, \bibinfo{author}{Mindt, M.R.}, \bibinfo{author}{Wolf, M.}, \bibinfo{author}{Curtis, L.M.}, \bibinfo{author}{Agarwal, P.}, \bibinfo{author}{Wisnivesky, J.}, \bibinfo{author}{Nadkarni, G.N.}, et~al., \bibinfo{year}{2022}.
\newblock \bibinfo{title}{Natural language processing to identify patients with cognitive impairment}.
\newblock \bibinfo{journal}{medRxiv} , \bibinfo{pages}{2022--02}.
\bibitem[{Ilias and Askounis(2022)}]{ilias2022multimodal}
\bibinfo{author}{Ilias, L.}, \bibinfo{author}{Askounis, D.}, \bibinfo{year}{2022}.
\newblock \bibinfo{title}{Multimodal deep learning models for detecting dementia from speech and transcripts}.
\newblock \bibinfo{journal}{Frontiers in Aging Neuroscience} \bibinfo{volume}{14}, \bibinfo{pages}{830943}.
\bibitem[{Kingma and Ba(2014)}]{kingma2014adam}
\bibinfo{author}{Kingma, D.P.}, \bibinfo{author}{Ba, J.}, \bibinfo{year}{2014}.
\newblock \bibinfo{title}{Adam: A method for stochastic optimization}.
\newblock \bibinfo{journal}{arXiv preprint arXiv:1412.6980} .
\bibitem[{Kullback and Leibler(1951)}]{kullback1951information}
\bibinfo{author}{Kullback, S.}, \bibinfo{author}{Leibler, R.A.}, \bibinfo{year}{1951}.
\newblock \bibinfo{title}{On information and sufficiency}.
\newblock \bibinfo{journal}{The annals of mathematical statistics} \bibinfo{volume}{22}, \bibinfo{pages}{79--86}.
\bibitem[{Langa and Levine(2014)}]{langa2014diagnosis}
\bibinfo{author}{Langa, K.M.}, \bibinfo{author}{Levine, D.A.}, \bibinfo{year}{2014}.
\newblock \bibinfo{title}{The diagnosis and management of mild cognitive impairment: a clinical review}.
\newblock \bibinfo{journal}{Jama} \bibinfo{volume}{312}, \bibinfo{pages}{2551--2561}.
\bibitem[{Lim et~al.(2021)Lim, Ar{\i}k, Loeff and Pfister}]{lim2021temporal}
\bibinfo{author}{Lim, B.}, \bibinfo{author}{Ar{\i}k, S.{\"O}.}, \bibinfo{author}{Loeff, N.}, \bibinfo{author}{Pfister, T.}, \bibinfo{year}{2021}.
\newblock \bibinfo{title}{Temporal fusion transformers for interpretable multi-horizon time series forecasting}.
\newblock \bibinfo{journal}{International Journal of Forecasting} \bibinfo{volume}{37}, \bibinfo{pages}{1748--1764}.
\bibitem[{Lo et~al.(2020)Lo, Wang, Neumann, Kinney and Weld}]{lo-etal-2020-s2orc}
\bibinfo{author}{Lo, K.}, \bibinfo{author}{Wang, L.L.}, \bibinfo{author}{Neumann, M.}, \bibinfo{author}{Kinney, R.}, \bibinfo{author}{Weld, D.}, \bibinfo{year}{2020}.
\newblock \bibinfo{title}{{S}2{ORC}: The semantic scholar open research corpus}, in: \bibinfo{booktitle}{Proceedings of the 58th Annual Meeting of the Association for Computational Linguistics}, \bibinfo{publisher}{Association for Computational Linguistics}, \bibinfo{address}{Online}. pp. \bibinfo{pages}{4969--4983}.
\newblock \URLprefix \url{https://aclanthology.org/2020.acl-main.447}, \DOIprefix\doi{10.18653/v1/2020.acl-main.447}.
\bibitem[{Luz et~al.(2020)Luz, Haider, de~la Fuente, Fromm and MacWhinney}]{luz2020alzheimer}
\bibinfo{author}{Luz, S.}, \bibinfo{author}{Haider, F.}, \bibinfo{author}{de~la Fuente, S.}, \bibinfo{author}{Fromm, D.}, \bibinfo{author}{MacWhinney, B.}, \bibinfo{year}{2020}.
\newblock \bibinfo{title}{Alzheimer's dementia recognition through spontaneous speech: The adress challenge}.
\newblock \bibinfo{journal}{arXiv preprint arXiv:2004.06833} .
\bibitem[{{lzheimer’s~Association}(2023)}]{fs}
\bibinfo{author}{{lzheimer’s~Association}}, \bibinfo{year}{2023}.
\newblock \bibinfo{title}{2023 alzheimer's disease facts and figures}.
\newblock \bibinfo{journal}{Alzheimer's \& Dementia} \bibinfo{volume}{19}, \bibinfo{pages}{1598--1695}.
\newblock \URLprefix \url{https://doi.org/10.1002%2Falz.13016}, \DOIprefix\doi{10.1002/alz.13016}.
\bibitem[{vd~Maaten and Hinton(2008)}]{vd2008visualizing}
\bibinfo{author}{vd~Maaten, L.}, \bibinfo{author}{Hinton, G.}, \bibinfo{year}{2008}.
\newblock \bibinfo{title}{Visualizing data using t-sne}.
\newblock \bibinfo{journal}{Journal of machine learning research} \bibinfo{volume}{9}, \bibinfo{pages}{2579--2605}.
\bibitem[{Pappagari et~al.(2021)Pappagari, Cho, Joshi, Moro-Vel{\'a}zquez, Zelasko, Villalba and Dehak}]{pappagari2021automatic}
\bibinfo{author}{Pappagari, R.}, \bibinfo{author}{Cho, J.}, \bibinfo{author}{Joshi, S.}, \bibinfo{author}{Moro-Vel{\'a}zquez, L.}, \bibinfo{author}{Zelasko, P.}, \bibinfo{author}{Villalba, J.}, \bibinfo{author}{Dehak, N.}, \bibinfo{year}{2021}.
\newblock \bibinfo{title}{Automatic detection and assessment of alzheimer disease using speech and language technologies in low-resource scenarios.}, in: \bibinfo{booktitle}{Interspeech}, pp. \bibinfo{pages}{3825--3829}.
\bibitem[{Park et~al.(2019)Park, Chan, Zhang, Chiu, Zoph, Cubuk and Le}]{park2019specaugment}
\bibinfo{author}{Park, D.S.}, \bibinfo{author}{Chan, W.}, \bibinfo{author}{Zhang, Y.}, \bibinfo{author}{Chiu, C.C.}, \bibinfo{author}{Zoph, B.}, \bibinfo{author}{Cubuk, E.D.}, \bibinfo{author}{Le, Q.V.}, \bibinfo{year}{2019}.
\newblock \bibinfo{title}{Specaugment: A simple data augmentation method for automatic speech recognition}.
\newblock \bibinfo{journal}{arXiv preprint arXiv:1904.08779} .
\bibitem[{Penfold et~al.(2022)Penfold, Carrell, Cronkite, Pabiniak, Dodd, Glass, Johnson, Thompson, Arrighi and Stang}]{penfold2022development}
\bibinfo{author}{Penfold, R.B.}, \bibinfo{author}{Carrell, D.S.}, \bibinfo{author}{Cronkite, D.J.}, \bibinfo{author}{Pabiniak, C.}, \bibinfo{author}{Dodd, T.}, \bibinfo{author}{Glass, A.M.}, \bibinfo{author}{Johnson, E.}, \bibinfo{author}{Thompson, E.}, \bibinfo{author}{Arrighi, H.M.}, \bibinfo{author}{Stang, P.E.}, \bibinfo{year}{2022}.
\newblock \bibinfo{title}{Development of a machine learning model to predict mild cognitive impairment using natural language processing in the absence of screening}.
\newblock \bibinfo{journal}{BMC Medical Informatics and Decision Making} \bibinfo{volume}{22}, \bibinfo{pages}{1--13}.
\bibitem[{Pennington et~al.(2014)Pennington, Socher and Manning}]{pennington2014glove}
\bibinfo{author}{Pennington, J.}, \bibinfo{author}{Socher, R.}, \bibinfo{author}{Manning, C.D.}, \bibinfo{year}{2014}.
\newblock \bibinfo{title}{Glove: Global vectors for word representation}, in: \bibinfo{booktitle}{Proceedings of the 2014 conference on empirical methods in natural language processing (EMNLP)}, pp. \bibinfo{pages}{1532--1543}.
\bibitem[{Petersen(2004)}]{petersen2004mild}
\bibinfo{author}{Petersen, R.C.}, \bibinfo{year}{2004}.
\newblock \bibinfo{title}{Mild cognitive impairment as a diagnostic entity}.
\newblock \bibinfo{journal}{Journal of internal medicine} \bibinfo{volume}{256}, \bibinfo{pages}{183--194}.
\bibitem[{Petersen et~al.(2014)Petersen, Caracciolo, Brayne, Gauthier, Jelic and Fratiglioni}]{Petersen2014}
\bibinfo{author}{Petersen, R.C.}, \bibinfo{author}{Caracciolo, B.}, \bibinfo{author}{Brayne, C.}, \bibinfo{author}{Gauthier, S.}, \bibinfo{author}{Jelic, V.}, \bibinfo{author}{Fratiglioni, L.}, \bibinfo{year}{2014}.
\newblock \bibinfo{title}{Mild cognitive impairment: a concept in evolution}.
\newblock \bibinfo{journal}{Journal of Internal Medicine} \bibinfo{volume}{275}, \bibinfo{pages}{214--228}.
\newblock \URLprefix \url{https://doi.org/10.1111/joim.12190}, \DOIprefix\doi{10.1111/joim.12190}.
\bibitem[{Pompili et~al.(2020)Pompili, Rolland and Abad}]{pompili2020inesc}
\bibinfo{author}{Pompili, A.}, \bibinfo{author}{Rolland, T.}, \bibinfo{author}{Abad, A.}, \bibinfo{year}{2020}.
\newblock \bibinfo{title}{The inesc-id multi-modal system for the adress 2020 challenge}.
\newblock \bibinfo{journal}{arXiv preprint arXiv:2005.14646} .
\bibitem[{Reimers and Gurevych(2019)}]{reimers2019sentence}
\bibinfo{author}{Reimers, N.}, \bibinfo{author}{Gurevych, I.}, \bibinfo{year}{2019}.
\newblock \bibinfo{title}{Sentence-bert: Sentence embeddings using siamese bert-networks}, in: \bibinfo{booktitle}{Proceedings of the 2019 Conference on Empirical Methods in Natural Language Processing and the 9th International Joint Conference on Natural Language Processing (EMNLP-IJCNLP)}, pp. \bibinfo{pages}{3982--3992}.
\bibitem[{Roark et~al.(2011)Roark, Mitchell, Hosom, Hollingshead and Kaye}]{roark2011spoken}
\bibinfo{author}{Roark, B.}, \bibinfo{author}{Mitchell, M.}, \bibinfo{author}{Hosom, J.P.}, \bibinfo{author}{Hollingshead, K.}, \bibinfo{author}{Kaye, J.}, \bibinfo{year}{2011}.
\newblock \bibinfo{title}{Spoken language derived measures for detecting mild cognitive impairment}.
\newblock \bibinfo{journal}{IEEE transactions on audio, speech, and language processing} \bibinfo{volume}{19}, \bibinfo{pages}{2081--2090}.
\bibitem[{Rohanian et~al.(2021)Rohanian, Hough and Purver}]{rohanian2021multi}
\bibinfo{author}{Rohanian, M.}, \bibinfo{author}{Hough, J.}, \bibinfo{author}{Purver, M.}, \bibinfo{year}{2021}.
\newblock \bibinfo{title}{Multi-modal fusion with gating using audio, lexical and disfluency features for alzheimer's dementia recognition from spontaneous speech}.
\newblock \bibinfo{journal}{arXiv preprint arXiv:2106.09668} .
\bibitem[{Roshanzamir et~al.(2021)Roshanzamir, Aghajan and Soleymani~Baghshah}]{roshanzamir2021transformer}
\bibinfo{author}{Roshanzamir, A.}, \bibinfo{author}{Aghajan, H.}, \bibinfo{author}{Soleymani~Baghshah, M.}, \bibinfo{year}{2021}.
\newblock \bibinfo{title}{Transformer-based deep neural network language models for alzheimer’s disease risk assessment from targeted speech}.
\newblock \bibinfo{journal}{BMC Medical Informatics and Decision Making} \bibinfo{volume}{21}, \bibinfo{pages}{1--14}.
\bibitem[{Santos et~al.(2017)Santos, Corr{\^e}a~Jr, Oliveira~Jr, Amancio, Mansur and Alu{\'\i}sio}]{santos2017enriching}
\bibinfo{author}{Santos, L.B.d.}, \bibinfo{author}{Corr{\^e}a~Jr, E.A.}, \bibinfo{author}{Oliveira~Jr, O.N.}, \bibinfo{author}{Amancio, D.R.}, \bibinfo{author}{Mansur, L.L.}, \bibinfo{author}{Alu{\'\i}sio, S.M.}, \bibinfo{year}{2017}.
\newblock \bibinfo{title}{Enriching complex networks with word embeddings for detecting mild cognitive impairment from speech transcripts}.
\newblock \bibinfo{journal}{arXiv preprint arXiv:1704.08088} .
\bibitem[{Song et~al.(2020)Song, Tan, Qin, Lu and Liu}]{song2020mpnet}
\bibinfo{author}{Song, K.}, \bibinfo{author}{Tan, X.}, \bibinfo{author}{Qin, T.}, \bibinfo{author}{Lu, J.}, \bibinfo{author}{Liu, T.Y.}, \bibinfo{year}{2020}.
\newblock \bibinfo{title}{Mpnet: Masked and permuted pre-training for language understanding}.
\newblock \bibinfo{journal}{Advances in Neural Information Processing Systems} \bibinfo{volume}{33}, \bibinfo{pages}{16857--16867}.
\bibitem[{Sperling et~al.(2011)Sperling, Aisen, Beckett, Bennett, Craft, Fagan, Iwatsubo, Jack~Jr, Kaye, Montine et~al.}]{sperling2011toward}
\bibinfo{author}{Sperling, R.A.}, \bibinfo{author}{Aisen, P.S.}, \bibinfo{author}{Beckett, L.A.}, \bibinfo{author}{Bennett, D.A.}, \bibinfo{author}{Craft, S.}, \bibinfo{author}{Fagan, A.M.}, \bibinfo{author}{Iwatsubo, T.}, \bibinfo{author}{Jack~Jr, C.R.}, \bibinfo{author}{Kaye, J.}, \bibinfo{author}{Montine, T.J.}, et~al., \bibinfo{year}{2011}.
\newblock \bibinfo{title}{Toward defining the preclinical stages of alzheimer’s disease: Recommendations from the national institute on aging-alzheimer's association workgroups on diagnostic guidelines for alzheimer's disease}.
\newblock \bibinfo{journal}{Alzheimer's \& dementia} \bibinfo{volume}{7}, \bibinfo{pages}{280--292}.
\bibitem[{Sun et~al.(2023)Sun, Dodge and Mahoor}]{sun2023mc}
\bibinfo{author}{Sun, J.}, \bibinfo{author}{Dodge, H.H.}, \bibinfo{author}{Mahoor, M.H.}, \bibinfo{year}{2023}.
\newblock \bibinfo{title}{Mc-vivit: Multi-branch classifier-vivit to detect mild cognitive impairment in older adults using facial videos}.
\newblock \bibinfo{journal}{arXiv preprint arXiv:2304.05292} .
\bibitem[{Syed et~al.(2021)Syed, Syed, Lech and Pirogova}]{syed2021automated}
\bibinfo{author}{Syed, Z.S.}, \bibinfo{author}{Syed, M.S.S.}, \bibinfo{author}{Lech, M.}, \bibinfo{author}{Pirogova, E.}, \bibinfo{year}{2021}.
\newblock \bibinfo{title}{Automated recognition of alzheimer’s dementia using bag-of-deep-features and model ensembling}.
\newblock \bibinfo{journal}{IEEE Access} \bibinfo{volume}{9}, \bibinfo{pages}{88377--88390}.
\bibitem[{Tang et~al.(2022)Tang, Chen, Dodge and Zhou}]{tang2022joint}
\bibinfo{author}{Tang, F.}, \bibinfo{author}{Chen, J.}, \bibinfo{author}{Dodge, H.H.}, \bibinfo{author}{Zhou, J.}, \bibinfo{year}{2022}.
\newblock \bibinfo{title}{The joint effects of acoustic and linguistic markers for early identification of mild cognitive impairment}.
\newblock \bibinfo{journal}{Frontiers in digital health} \bibinfo{volume}{3}, \bibinfo{pages}{702772}.
\bibitem[{Toledo et~al.(2018)Toledo, Alu{\'\i}sio, Dos~Santos, Brucki, Tr{\'e}s, de~Oliveira and Mansur}]{toledo2018analysis}
\bibinfo{author}{Toledo, C.M.}, \bibinfo{author}{Alu{\'\i}sio, S.M.}, \bibinfo{author}{Dos~Santos, L.B.}, \bibinfo{author}{Brucki, S.M.D.}, \bibinfo{author}{Tr{\'e}s, E.S.}, \bibinfo{author}{de~Oliveira, M.O.}, \bibinfo{author}{Mansur, L.L.}, \bibinfo{year}{2018}.
\newblock \bibinfo{title}{Analysis of macrolinguistic aspects of narratives from individuals with alzheimer's disease, mild cognitive impairment, and no cognitive impairment}.
\newblock \bibinfo{journal}{Alzheimer's \& Dementia: Diagnosis, Assessment \& Disease Monitoring} \bibinfo{volume}{10}, \bibinfo{pages}{31--40}.
\bibitem[{Vaswani et~al.(2017)Vaswani, Shazeer, Parmar, Uszkoreit, Jones, Gomez, Kaiser and Polosukhin}]{vaswani2017attention}
\bibinfo{author}{Vaswani, A.}, \bibinfo{author}{Shazeer, N.}, \bibinfo{author}{Parmar, N.}, \bibinfo{author}{Uszkoreit, J.}, \bibinfo{author}{Jones, L.}, \bibinfo{author}{Gomez, A.N.}, \bibinfo{author}{Kaiser, {\L}.}, \bibinfo{author}{Polosukhin, I.}, \bibinfo{year}{2017}.
\newblock \bibinfo{title}{Attention is all you need}.
\newblock \bibinfo{journal}{Advances in neural information processing systems} \bibinfo{volume}{30}.
\bibitem[{Xu et~al.(2019)Xu, Meng, Qiu, Yu and Wu}]{xu2019sentiment}
\bibinfo{author}{Xu, G.}, \bibinfo{author}{Meng, Y.}, \bibinfo{author}{Qiu, X.}, \bibinfo{author}{Yu, Z.}, \bibinfo{author}{Wu, X.}, \bibinfo{year}{2019}.
\newblock \bibinfo{title}{Sentiment analysis of comment texts based on bilstm}.
\newblock \bibinfo{journal}{Ieee Access} \bibinfo{volume}{7}, \bibinfo{pages}{51522--51532}.
\bibitem[{Yamada et~al.(2022)Yamada, Shinkawa, Nemoto, Ota, Nemoto and Arai}]{yamada2022speech}
\bibinfo{author}{Yamada, Y.}, \bibinfo{author}{Shinkawa, K.}, \bibinfo{author}{Nemoto, M.}, \bibinfo{author}{Ota, M.}, \bibinfo{author}{Nemoto, K.}, \bibinfo{author}{Arai, T.}, \bibinfo{year}{2022}.
\newblock \bibinfo{title}{Speech and language characteristics differentiate alzheimer's disease and dementia with lewy bodies}.
\newblock \bibinfo{journal}{Alzheimer's \& Dementia: Diagnosis, Assessment \& Disease Monitoring} \bibinfo{volume}{14}, \bibinfo{pages}{e12364}.
\bibitem[{Yeung et~al.(2021)Yeung, Iaboni, Rochon, Lavoie, Santiago, Yancheva, Novikova, Xu, Robin, Kaufman et~al.}]{yeung2021correlating}
\bibinfo{author}{Yeung, A.}, \bibinfo{author}{Iaboni, A.}, \bibinfo{author}{Rochon, E.}, \bibinfo{author}{Lavoie, M.}, \bibinfo{author}{Santiago, C.}, \bibinfo{author}{Yancheva, M.}, \bibinfo{author}{Novikova, J.}, \bibinfo{author}{Xu, M.}, \bibinfo{author}{Robin, J.}, \bibinfo{author}{Kaufman, L.D.}, et~al., \bibinfo{year}{2021}.
\newblock \bibinfo{title}{Correlating natural language processing and automated speech analysis with clinician assessment to quantify speech-language changes in mild cognitive impairment and alzheimer’s dementia}.
\newblock \bibinfo{journal}{Alzheimer's research \& therapy} \bibinfo{volume}{13}, \bibinfo{pages}{109}.
\bibitem[{Yu et~al.(2021)Yu, Wild, Potempa, Hampstead, Lichtenberg, Struble, Pruitt, Alfaro, Lindsley, MacDonald et~al.}]{yu2021internet}
\bibinfo{author}{Yu, K.}, \bibinfo{author}{Wild, K.}, \bibinfo{author}{Potempa, K.}, \bibinfo{author}{Hampstead, B.M.}, \bibinfo{author}{Lichtenberg, P.A.}, \bibinfo{author}{Struble, L.M.}, \bibinfo{author}{Pruitt, P.}, \bibinfo{author}{Alfaro, E.L.}, \bibinfo{author}{Lindsley, J.}, \bibinfo{author}{MacDonald, M.}, et~al., \bibinfo{year}{2021}.
\newblock \bibinfo{title}{The internet-based conversational engagement clinical trial (i-conect) in socially isolated adults 75+ years old: randomized controlled trial protocol and covid-19 related study modifications}.
\newblock \bibinfo{journal}{Frontiers in digital health} \bibinfo{volume}{3}, \bibinfo{pages}{714813}.
\bibitem[{Yuan et~al.(2021)Yuan, Cai, Bian, Ye and Church}]{yuan2021pauses}
\bibinfo{author}{Yuan, J.}, \bibinfo{author}{Cai, X.}, \bibinfo{author}{Bian, Y.}, \bibinfo{author}{Ye, Z.}, \bibinfo{author}{Church, K.}, \bibinfo{year}{2021}.
\newblock \bibinfo{title}{Pauses for detection of alzheimer’s disease}.
\newblock \bibinfo{journal}{Frontiers in Computer Science} \bibinfo{volume}{2}, \bibinfo{pages}{624488}.
\bibitem[{Zeng et~al.(2023)Zeng, Chen, Zhang and Xu}]{zeng2023transformers}
\bibinfo{author}{Zeng, A.}, \bibinfo{author}{Chen, M.}, \bibinfo{author}{Zhang, L.}, \bibinfo{author}{Xu, Q.}, \bibinfo{year}{2023}.
\newblock \bibinfo{title}{Are transformers effective for time series forecasting?}, in: \bibinfo{booktitle}{Proceedings of the AAAI conference on artificial intelligence}, pp. \bibinfo{pages}{11121--11128}.
\bibitem[{Zhou et~al.(2021)Zhou, Zhang, Peng, Zhang, Li, Xiong and Zhang}]{zhou2021informer}
\bibinfo{author}{Zhou, H.}, \bibinfo{author}{Zhang, S.}, \bibinfo{author}{Peng, J.}, \bibinfo{author}{Zhang, S.}, \bibinfo{author}{Li, J.}, \bibinfo{author}{Xiong, H.}, \bibinfo{author}{Zhang, W.}, \bibinfo{year}{2021}.
\newblock \bibinfo{title}{Informer: Beyond efficient transformer for long sequence time-series forecasting}, in: \bibinfo{booktitle}{Proceedings of the AAAI conference on artificial intelligence}, pp. \bibinfo{pages}{11106--11115}.
\bibitem[{Zolnoori et~al.(2023)Zolnoori, Zolnour and Topaz}]{zolnoori2023adscreen}
\bibinfo{author}{Zolnoori, M.}, \bibinfo{author}{Zolnour, A.}, \bibinfo{author}{Topaz, M.}, \bibinfo{year}{2023}.
\newblock \bibinfo{title}{Adscreen: A speech processing-based screening system for automatic identification of patients with alzheimer's disease and related dementia}.
\newblock \bibinfo{journal}{Artificial Intelligence in Medicine} \bibinfo{volume}{143}, \bibinfo{pages}{102624}.

\end{thebibliography}

\end{document}